# Modelling and simulation of a commercially available dielectric elastomer actuator

Lukas Sohlbach, Hamza Hobbani, Chistopher Blase, Fernando Perez-Peña and Karsten Schmidt

*Abstract*—**In order to fully harness the potential of dielectric elastomer actuators (DEAs) in soft robots, advanced control methods are needed. An important groundwork for this is the development of a control-oriented model that can adequately describe the underlying dynamics of a DEA. A common feature of existing models is that always custom-made DEAs were investigated. This makes the modelling process easier, as all specifications and the structure of the actuator are well known. In the case of a commercial actuator, however, only the information from the manufacturer is available and must be checked or completed during the modelling process. The aim of this paper is to explore how a commercial stacked silicone-based DEA can be modelled and how complex the model should be to properly replicate the features of the actuator. The static description has demonstrated the suitability of Hooke's law. In the case of dynamic description, it is shown that no viscoelastic model is needed for control-oriented modelling. However, if all features of the DEA are considered, the generalized Kelvin-Maxwell model with three Maxwell elements shows good results, stability and computational efficiency.**

*Index Terms*— **Modeling, Control, and Learning for Soft Robots; Soft Robot Materials and Design; Soft Sensors and Actuators Dielectric elastomer actuators**

## I. INTRODUCTION

Dielectric elastomer actuators (DEAs), as a group of artificial muscles [1], belong to the core components of soft robots creating robotic systems that are safer, cheaper and more flexible than traditional ones [2, 3]. Due to the elastomers used, the dynamics of DEAs exhibit non-linear and time-dependent effects [4, 5]. In order to exploit their full potential, advanced control methods are needed [6, 7].

Manuscript received Month dd, yyyy; revised Month dd, yyyy. This paper was funded and supported by the Frankfurt University of Applied Sciences and the University of Cadiz. The authors acknowledge the valuable support of Prof. Dr.-Ing. Stefan Dominico (Frankfurt University of Applied Sciences) and CTsystems AG. (*Corresponding author: Lukas Sohlbach)*.

Lukas Sohlbach and Hamza Hobbani are with Department of Computer Science and Engineering, Frankfurt University of Applied Sciences, Frankfurt/Main, Germany and the Department of Automation, Electronics and Computer Architecture and Networks, School of Engineering, University of Cadiz, Cadiz, Spain (e-mail: lukas.sohlbach@fb2.fra-uas.de; h.hobbani@gmail.com)

Christopher Blase and Karsten Schmidt are with the Department of Computer Science and Engineering, Frankfurt University of Applied Sciences, Frankfurt/Main, Germany (e-mail: cblase@fb2.fra-uas.de; schmidtk@fb2.fra-uas.de)

Fernando Perez-Peña is with the Department of Automation, Electronics and Computer Architecture and Networks, School of Engineering, University of Cadiz, Cadiz, Spain (e-mail: fernandoperez.pena@uca.es).

An important groundwork is the development of a control-oriented model to adequately describe the underlying dynamics of a DEA [7].

Different methods of system identification have been applied to describe the active and the passive part of a DEA, leading to a mathematical relationship between the input (voltage) and output (deformation) variable [8]. When referring to the active part, the electrostatic force is meant, which actively compresses the DEA. By passive part, one refers to the elastomer, since the material passively pushes the DEA back in the direction of the initial configuration. Most of the time, the results are so-called grey-box models, where the model structure for the active part is based on Maxwell stress [9] (also known as Maxwell pressure or effective pressure). The passive part is described using statistical or continuum mechanics. Parameters of the respective structure are determined partly or completely by experiments [8, 10, 11]. Statistical mechanics are based on Gaussian and non-Gaussian statistics and attempt to explain relationships microscopically [10, 12]. In contrast, continuum mechanics, as a phenomenological approach, attempt to describe the deformation behavior of a solid, gas or fluid through macroscopic observations [13, 14]. In the context of DEAs, statistical mechanics model structures are used to describe the material statically, while continuum mechanics structures can be used to describe both the material statically and dynamically. In some cases, so-called black-box models are also used. With these models, the focus is on the representation of the data, regardless of the mathematical structure of the model [8].

In [15] a model for a silicone based stacked DEA was developed. The electrostatic force was represented with the equation proposed by Pelrine et al. [9]. For the static description of the material the Hookean law was used. A generalized Maxwell was employed to reproduce the dynamics. To implement the model varying parameters were used. In addition to the Hookean law, which is also used in [6, 16, 17], other material models were also applied to describe static behavior (Gent [18] and [19] acrylic-based membrane DEAs, Yeoh [20] and [21] acrylic [20] and PU [21] stacked DEA). All four papers described the electrostatic force with [9] and the dynamics with a generalized Maxwell model. They implemented the models with one set of parameters. A smaller part of the reviewed works also used the generalized Kelvin-Voight model to represent the dynamics of acrylic and silicone based membrane DEAs [22–24]. The equation from [9] was used for the electrostatic force. In [24] the Neo-Hookean model was used for the static description but no precise information was given in [22, 23].

Both used one set of parameters to implement the model. Jacobs et al. [7] and Cao et al. [25] identified black-box models to describe their acrylic based membrane DEAs and used a NARX (nonlinear autoregressive with exogenous input) and a transfer function as model structure. Both models were implemented with one set of parameters.

The majority of the papers discussed used MATLAB/Simulink to identify the model parameters and implement or simulate the model. FEM (finite element method) programs such as ANSYS [15] or COSMOL [20] were also used in some cases or no software was specified [7, 18, 19, 22, 24]. Among the cited studies, there are different approaches to the validation of the models. In [15], only the passive part of the model was validated using pure elastomer experiments. Works that had the control of the DEAs as their focus only validated the controller developed on the basis of the model [16, 23, 24]. Others validated their model with data contained in the identification data [18, 19]. In the remaining papers analyzed, there seems to be some ambiguity about the validation process and the inclusion of amplitudes or frequencies of identification. [6, 7, 20–22, 25]. A common feature of all the articles is that custom-made DEAs were modelled. This makes the modelling process easier, as all specifications and the structure of the actuator are well known. In the case of a commercial actuator, however, only the information from the manufacturer is available and this might not include all the details needed to build a theoretical model. The objective of this paper is to explore how a commercial stacked silicone-based DEA can be modelled. In addition, it is shown how complex the model has to be to meet the experimental data. In contrast to the works mentioned above, mechanical restrictions such as fixed end plates are also characterized and the computational efficiency of the models is considered. For this purpose, both linear elastic and hyperealstic approaches are investigated for static passive modelling. Subsequently, these are extended in terms of dynamics with common viscoelastic models and by an approach from system identification.

## II. Dielectric Elastomer Actuators

The simplest configuration of a DEA can be seen as a flexible capacitor [26]. An incompressible elastomer that is sandwiched between two compliant electrodes and deforms when an electric field is applied across the electrodes through a voltage [26, 27]. Figure 1 a) displays the DEA in reference state, whereas Figure 1 b) shows the actuated state.

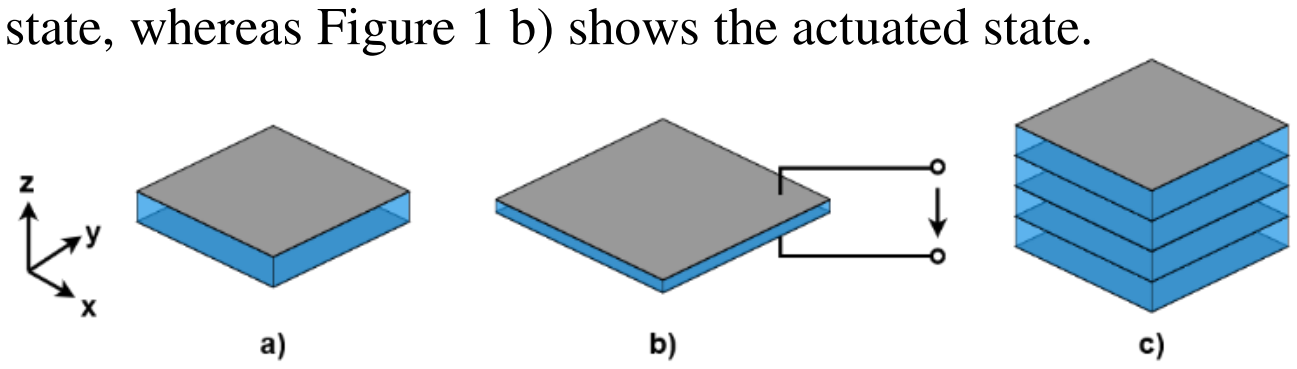

**Fig. 1.** Dielectric elastomer actuator a) not activated b) activated c) stacked.

The DEA's working principle is based on the electrostatic concept. A Coulomb force is generated in response to the applied voltage.

The force induces a stress in the dielectric elastomer (DE) membrane, which results in a thickness reduction [26]. Using (1), based on the approach of Pelrine et al. [9], the stress inside the DE membrane can be calculated.

$$\sigma_{el}(t) = \varepsilon_0 \cdot \varepsilon_r \cdot E(t)^2 = \varepsilon_0 \cdot \varepsilon_r \cdot \left(\frac{V(t)}{z}\right)^2 \qquad (1)$$

With $\varepsilon_0$ and $\varepsilon_r$ as absolute and relative permittivity of the DE membrane, $E$ as the electric field, $V(t)$ as the applied voltage and $z$ the thickness of the DE membrane in its non-actuated state. In contrast to [9], a variable voltage is considered, which leads to a temporal dependency for the equation and also for the following static consideration. For (1), it is assumed that the charges are evenly distributed on infinitely sized electrodes and the membrane has a uniform thickness. In addition that the dielectric properties of the elastomer do not change due to deformation (ideal elastomer) [26]. Equation 1 is only valid for deformations up to 10 %, beyond that, the change in membrane thickness must be considered. The membrane of a DEA is compressed by the electrostatic pressure until this pressure and the stress inside the elastomer are in equilibrium [28]. Looking at the DE membrane in the resting state at the molecular level, twisted polymer chains can be found, as illustrated in Figure 2 a) [29].

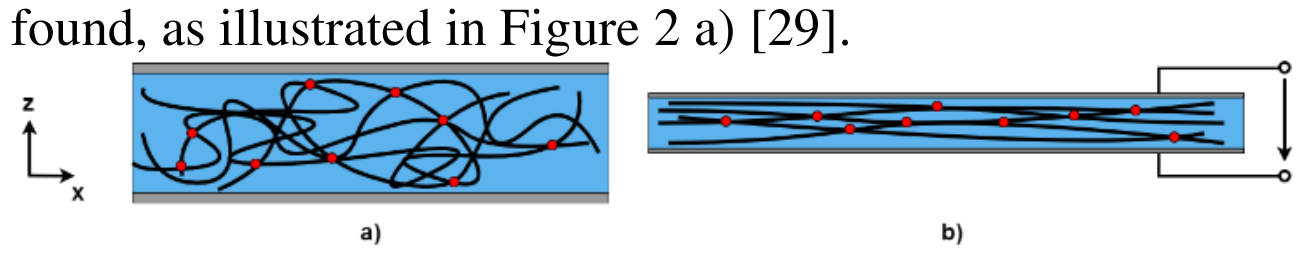

**Fig. 2.** Polymer chains in the a) reference b) actuated state.

The red dots represent so-called crosslinks, which limit the mobility of the individual chains. When a force is applied to the elastomer, the polymer chains reconfigure themselves to distribute the stress as displayed in Figure 2b) [29]. The unfolding of the polymer chains is also responsible for the fact that the stress-strain curve, especially at high strains, shows a non-linear behavior [30]. As schematically shown in Figure 3 a), the stress-strain curve initially has a linear range (up to 10% deformation). As soon as the rearrangement of the polymer chains takes place, the non-linear range begins. When the chains reach their finite length, the material stiffens and the stress increases rapidly until failure occurs [26]. It should be noted that the deformation described does not happen without dissipation [26]. It leads to time-dependent effects such as hysteresis, creep and relaxation as schematically shown in Figure 3 b)-d) (for c) and d) stimulus in the upper and response in the lower part) [11, 26].

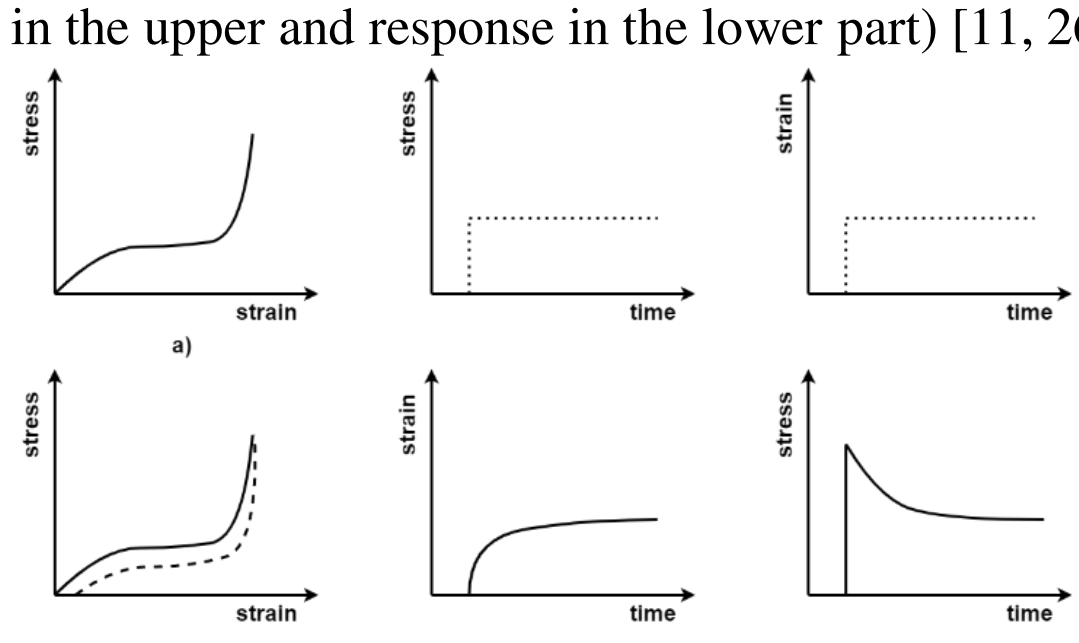

**Fig. 3.** Schematic elastomer deformation curves a) static stress-strain b) hysteresis c) creep d) relaxation

Various material models are available to describe the properties of elastomers. The linear part of the stress-strain curve, for example, can be described using the linear elastic Hookean law. In addition, hyperelastic material models can be applied to represent the non-linear part [31]. To describe the time-dependent properties such as creep and relaxation, viscoelastic models such as the generalized Maxwell or the generalized Kelvin-Voigt model can be utilized. The generalized Maxwell model, on the one hand, is better suited to describe the decrease in stress during a constant deformation, hence the relaxation. The generalized Kelvin-Voigt model, on the other hand, is better suited to describe the increase in strain as a reaction to a constant stress, namely creep [32].

## III. Methodology

During this study, a commercially available stacked DEA from CTSystems (Ref. CT25.2-10-10 x 10 mm) was used as the object of investigation. In this type of actuator, silicone was used as membrane material and carbon powder as material for the electrodes and feeding lines [33]. The fundamental structure of a stacked actuator is shown in Figure 1 c). The individual layers are connected mechanically in series and electrically in parallel [20]. A special feature of the actuator from CTSystems is that the individual layers are first stacked into modules, which are tested and then stacked to the final actuator height, which is therefore always a multiple of the number of layers respectively modules [33]. The actuator height was specified by the manufacturer with 10 mm and the quadratic footprint with 10 mm by 10 mm [33, 34]. At the upper and lower end, the actuator was equipped with rigid end plates. These can be used to connect the actuator to its environment. They also prevent uneven deformation of the actuator's surface. The DEA has a maximum operating frequency of 100 Hz.

During modelling, the actuator was considered unloaded and it was assumed that there is a uniaxial state of stress in the main direction of motion (Z direction according to Figure 1). Furthermore, the actuator was treated as a unit. This means that the characterization was carried out for the combination of membrane and electrode material. Thus, the influence of mechanical boundary conditions such as the rigid end plates was to be characterized as well. Furthermore, the assumption was made that the behavior of the entire actuator can be described by the behavior of a single layer [21].

### A. Static modelling

As already mentioned in the introduction, the deformation of the actuator relies on an equilibrium of forces between the active and the passive part. With reference to the coordinate system in Figure 1, the following equilibrium of forces in the Z direction was considered for the static description.

$$\sum F_z \uparrow: \; -F_{active}(t) + F_{passive}(t) = 0 \quad (2)$$
$$with\; F_{active}(t) = \sigma_{el}(t) \cdot A_e$$
$$and\; F_{passive}(t) = \sigma_{z_{static}}(t) \cdot A_c$$

Where $A_e$ represents the area of the electrodes and $A_c$ the total area. For the dynamic description, the equilibrium of forces was extended by the inertia force, as shown in (3).

$$\sum F_z \uparrow: \; -F_{active}(t) + F_{passive}(t) + F_{In}(t) = 0 \quad (3)$$
$$with\; F_{active}(t) = \sigma_{el}(t) \cdot A_e,$$
$$F_{passive}(t) = A_c \cdot \left(\sigma_{z_{static}}(t) + \sigma_{z_{dynamic}}(t)\right),$$
$$and\; F_{in}(t) = m \cdot z_o \cdot \frac{\mathrm{d}^2 \varepsilon_z(t)}{\mathrm{d}t^2}$$

Equation 1 was used to calculate $\sigma_{el}(t)$. The change in thickness of the membrane during deformation was not considered. The following models were considered for the passive static description: Hookean, Neo-Hookean [35], Yeoh [36], Gent [37], Ogden [38] and Mooney-Rivlin [11, 39–42]. For reasons of clarity, the static description does not contain the visualization of the temporal progression. Appling the one dimensional Hookean law, the stress was related to the strain via a constant elastic modulus or Young's modulus, as shown in (4) [15].

$$\sigma_{z_{static}} = Y \cdot \varepsilon_z = Y \cdot (\lambda_z - 1) \quad (4)$$

With $\sigma_{z_{static}}$ as the true (Cauchy) stress in Z direction, $Y$ the modulus of elasticity, $\varepsilon_z = \frac{\Delta z}{z_0}$ the strain and $\lambda_z = \frac{z}{z_0}$ as the stretch ratio in Z direction. In contrast, for hyperelastic models, the relationship between stress and deformation was established via a strain energy function. For a uniaxial-loading condition (hydrostatic pressure p = 0), the relationship is shown in (5) [21].

$$\sigma_{z_{static}} = \lambda_z \cdot \frac{\partial W_s}{\partial \lambda_z} \quad (5)$$

With $\sigma_{z_{static}}$ as the true (Cauchy) stress and $\lambda_z$ as the stretch ratio in Z direction. The strain energy function describes the change in the free energy density according to the work done on the system and was initially adopted by Sou [27] to describe DEAs. His approach was based on a modification of the first law of thermodynamics and assumes that the temperature does not change during the operation of a DEA, which leads to the free energy as an isothermal invariant of the internal energy [27, 43]. The Table SI in the supplementary material contains detailed information on the hyperelastic models used. It should be noted that for the strain-energy function and the true stress, the first two main invariants of the Cauchy-Green deformation tensor were reduced using the incompressibility assumption of the elastomer ($\lambda_x \cdot \lambda_y \cdot \lambda_z = 1$) and the assumption that the deformation is unconfined, which means no restrictions and therefore no stresses in the X and Y direction ($\lambda_x = \lambda_y$).

### B. Dynamic modelling

In this paper, the passive static description was supplemented by the following passive dynamic models or model structures: generalized Maxwell (GM), generalized Kelvin-Voigt (GKV), generalized Kelvin-Maxwell (GKM) and a state space model (StSp). The Kelvin-Maxwell model is a combination of the Maxwell and Kelvin-Voigt models.

They are linear viscoelastic models that try to describe the macroscopic time-dependent material deformation by a combination of (hyper)elastic and viscous elements (springs and dampers). Figure 4 displays the mechanical equivalent diagrams of the models.

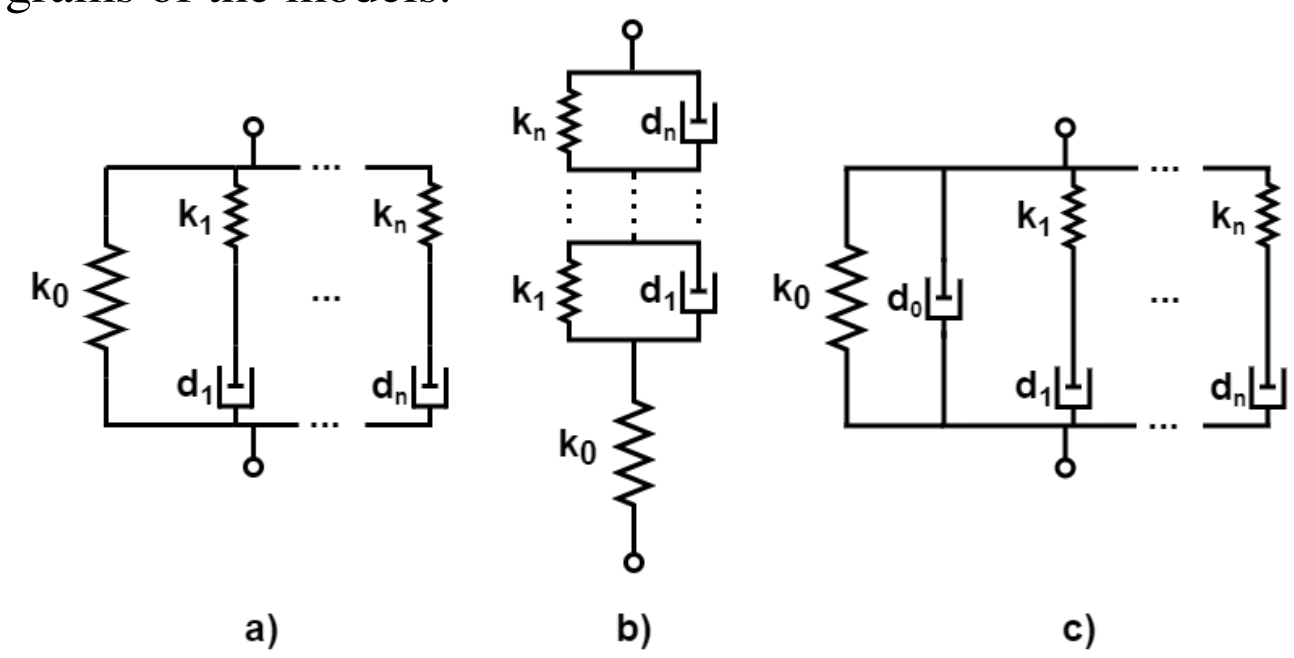

**Fig. 4.** Viscoelastic models a) generalized Maxwell b) generalized Kelvin-Voigt c) generalized Kelvin-Maxwell

As shown in Figure 4 a), the generalized Maxwell model consists of a spring and $n$ Maxwell elements arranged in parallel, which themselves are series connected springs and dampers. The spring $k_0$ represents the static deformation or the deformation of the material at time $t = \infty$. Via the parallel elements $k_n$ and $d_n$, the viscous losses can be represented [21]. By increasing the number of Maxwell elements, the frequency range that the model can represent can be increased [44]. The stress and deformation of the Maxwell elements is described by a differential equation. It is constructed from the equation of the mechanical components and is based on the approach that within a Maxwell element, the stress is the same for all components, but the strain is different [45]. In (6) the resulting differential equation is given.

$$\frac{\mathrm{d}\sigma_{MW}(t)}{\mathrm{d}t} = k \cdot \frac{\mathrm{d}\varepsilon_z(t)}{\mathrm{d}t} - \sigma_{MW}(t) \cdot \frac{k}{d} \quad (6)$$
$$with\ \sigma_{MW}(t) = \sigma_{spring}(t) = \sigma_{damper}(t)$$
$$and\ \varepsilon_z(t) = \varepsilon_{spring}(t) + \varepsilon_{damper}(t)$$

Where $\sigma_{MW}(t)$ represents the stress and $\varepsilon_z(t)$ the strain of one Maxwell element. If a constant strain is applied to a Maxwell element, the change in strain becomes zero $\left(\frac{\mathrm{d}\varepsilon_z(t)}{\mathrm{d}t} = 0\right)$ and (6) reduces to (7). Integration from 0 to t leads to (8) [46].

$$\frac{\mathrm{d}\sigma_{MW_{const}}(t)}{\mathrm{d}t} = -\sigma_{MW_{const}}(t) \cdot \frac{k}{d} \quad (7)$$
$$\sigma_{MW_{const}}(t) = \sigma_0 \cdot e^{-\frac{k \cdot t}{d}} \quad (8)$$

With $\sigma_0$ as the integration constant representing the peak stress after an applied strain step function. In contrast to the Maxwell model, the determination of the material response at time $t = \infty$ is not as straightforward for the Kelvin-Voigt model in Figure 4 b). It consists of a spring connected in series with Kelvin-Voigt elements, which in turn are parallel arrangements of springs and dampers.

The spring $k_0$ represents the spontaneous elastic reaction of the material [47] and the deformation at $t = \infty$ results from a combination of all springs. In addition, the elements $k_n$, together with the elements $d_n$, describe the creep process. By the number of Kelvin-Voigt elements, the frequency or time range to be represented is determined. When determining the Kelvin-Voigt differential equation, the approach is that the strain of the mechanical elements is the same, but the stress is different. This leads to the (9) for one Kelvin-Voigt element [45].

$$\sigma_{KV}(t) = k \cdot \varepsilon_z(t) + d \cdot \frac{\mathrm{d}\varepsilon_z(t)}{\mathrm{d}t} \quad (9)$$
$$with\ \varepsilon_z(t) = \varepsilon_{spring}(t) = \varepsilon_{damper}(t)$$
$$and\ \sigma_{KV}(t) = \sigma_{spring}(t) + \sigma_{damper}(t)$$

The stress is therefore a summation of the stresses of the mechanical elements. If several Kelvin-Voigt models or other elements are connected in series, the Maxwell element approach applies. For a constant stress this results in (10) [45]. If $t = \infty$, (10) reduces to (11).

$$\varepsilon_{z_{const}}(t) = \frac{\sigma_{KV_{const}}}{k_0} + \sum_{i=1}^{n} \frac{\sigma_{KV_{const}}}{k_i} \cdot \left(1 - e^{-\alpha_i \cdot t}\right) \quad (10)$$
$$with\ \alpha_i = \frac{k_i}{d_i}$$
$$\varepsilon_{z_{const}}(t = \infty) = \sum_{i=0}^{n} \frac{\sigma_{KV_{const}}}{k_i} \quad (11)$$

With $n$ as the number of Kelvin-Voigt elements, $\sigma_{KV_{const}}$ as the constant stress and $\varepsilon_{z_{const}}(t)$ as the strain. Equation 10 demonstrates that the deformation at time $t = \infty$ depends on the combination of all springs.

In this work, the aforementioned viscoelastic models were used for grey-box identification. The state space representation, however, was used as a black box approach. From a mathematical point of view, the state space representation is a set of $n$ fist-order differential equations written in vector notation. Thus, every linear differential equation of $n$-th order can be represented [48]. In the context of this work, a discrete third-order state space model, with no disturbance, was used to represent the dynamics of the system, since maximally a third-order differential tail equation results for the viscoelastic models in Figure 4. The resulting parameters do not have any physical meaning so far. In Table I the four applied dynamic models are given. Table SII in the supplementary material presents their corresponding passive forces or dynamic equations. It should be mentioned that $\sigma_z(t)$ for the second model already represents the combination of static and dynamic material model.

TABLE I
DYNAMIC MODELS

| No. | Static description | Dynamic description | Reference |
|---|---|---|---|
| 1 | Hookean law | Generalized Maxwell with n=3 | [21] |
| 2 | Hookean law | Generalized Kelvin-Voigt with n=3 | [49] |
| 3 | Hookean law | Generalized Kelvin-Maxwell with n=3 | [20] |
| 5 | Hookean law | Discrete 3rd order state space without disturbance | [50] |

For all viscoelastic models, the Hookean law was chosen as the static description because it is the simplest representation and thus the complexity and computational power of the models were to be reduced.

*C. Static identification*

In order to identify the parameters of the Hookean and the hyperelastic models from Table SI, a quasi-static compression test was carried out. For this purpose, the DEA was clamped between a linear actuator (Bose ElectroForce LM1 TestBench) and a fixed force sensor (ME KD24s 20N). The load cell was connected to the PCI signal input conditioner of the TestBench setup. The gain of the amplifier was adjusted by using a calibration resistor (6,8 Ω). The sensors' calibration was performed with a set of calibrated weights by adjusting the gain in the WinTest (7.01) software. In the test setup, the Z direction was in the horizontal direction and coincided with the moving direction of the linear actuator. Two adapters were developed for clamping the DEA, which had a 1 mm nose so that the DEA was supported on its rigid end plates and was barely compressed in the starting position. Figure 5 shows the DEA clamped in horizontal direction between the linear actuator and force sensor.

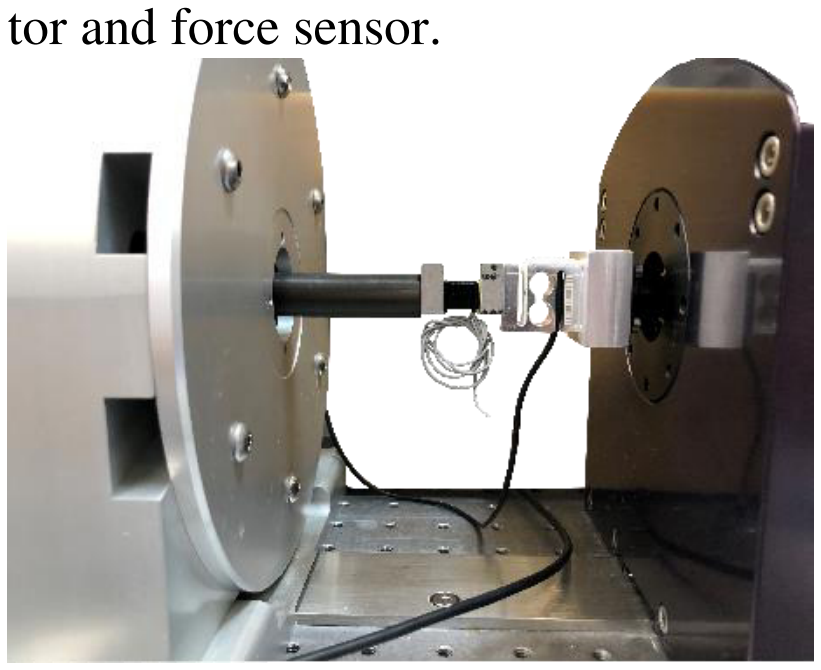

**Fig. 5.** Compression test setup.

In the experiment, the DEA was compressed by 20 % with a compression rate of 0.01 mm/s (2.5 mm in total). Using low compression speed was intended to reduce the viscoelastic effects [21]. The experiment included three cycles, which consisted of a loading and unloading phase and were carried out subsequently without a break. In this way, the material should be conditioned, as the material parameters for one-time and recurring loads differ due to the viscoelasticity and the parameters for recurring loads are interesting for an application-oriented simulation. WinTest was used to control the experiment and record the data: time, position of the linear actuator and force. A frequency of 200 Hz was used for recording and the position and force values were collected with an accuracy 5 μm and 0.02 N and a resolution of 0.1 μm and 0.001 N respectively.

After the experiment, the data was processed with MATLAB. In the beginning, the individual cycles were separated and only the third cycle was used for further processing. In order to calculate the actual initial uncompressed length of the DEA in relation to the position of the linear actuator, the first 10 % of the loading path were used to fit a linear equation using the MATLAB function polyfit. This fit was then used to calculate the actuator position for a force of 0 N. The calculated initial length was used to remove the initial range from measurement data. Furthermore, it was used to calculate the compressed DEA height $z$ from the remaining position data. Together with the uncompressed height of the elastomer stack $z_0$ (DEA height without the rigid end plates), which was measured with a profile projector (Keyence IM-750 and IM-6025) contactless (10.4120 mm), the stretch ratio $\lambda_z$ was calculated from the change in height. The profile projector has an accuracy of 5 $\mu$m and a resolution of 0.1 $\mu$m and was also used to measure the complete actuator area $A_c$ (11.429 mm by 10.018 mm). This area was then used to calculate the engineering stress from the measured force values. By multiplying the engineering stress with the stretch ratio, the true stress (Cauchy stress) was calculated [51]. It is worth mentioning that due to the equilibrium of forces in (2), the resulting stress is considered to be positive, although by convention the compressive stress occurring in the experiment should be considered negative. The relationship between true stress and strain ratio was used, utilizing the MATLAB function fit and the equations for $\sigma_{z_{static}}$ of Table SI, to identify the respective parameters. Within the fit function, the Nonlinear Least Squares method in combination with the Trust-Region algorithm was used. The maximum number of iterations was defined as 10000 and the function tolerance as $10^{-6}$.

*C. Dynamic identification*

The setup of the compression experiment was also used to determine the parameters of the model 1 (GM) from Table SII in a relaxation experiment. During the experiment, the DEA was compressed by 15 % for 300 seconds with a step function (1.75 mm in total) and the compression was held constant by the setup. A compression speed of 5 mm/s was used and the data should be collected according to Figure 3 d). The time, position of the linear actuator and the force were recorded with a frequency of 100 Hz. Accuracy and resolution correspond to the quasi-static experiment. During the relaxation test, no explicit preconditioning was carried out, as the previously executed compression tests served as preconditioning. After recording, the data was processed in a similar way to the compression experiment. In this experiment, however, the data was not trimmed with an initial length. For further processing, only the data from the stress peak to the point before the end of compression was used. In addition, using the Hookean law, the remaining elastic stress was calculated and subtracted from the stress of the experiment. Thus, only the stress seen by the three Maxwell elements is calculated [21].

Therefore, the value identified from the compression tests was used. For three elements, (8) leads to (12), which was subsequently used to identify the parameters.

$$\sigma_{MW_{const}}(t) = \sum_{i=1}^{3} \sigma_{MW_{const}}(t)_i = \varepsilon_0 \cdot k_1 \cdot e^{-\frac{k_1 \cdot t}{d_1}} + \varepsilon_0 \cdot k_2 \cdot e^{-\frac{k_2 \cdot t}{d_2}} + \varepsilon_0 \cdot k_3 \cdot e^{-\frac{k_3 \cdot t}{d_3}} \quad (12)$$

In this relationship $\varepsilon_0$ represents the applied constant compression. To perform the fitting, the MATLAB function fit with the Nonlinear Least Squares method was used. The maximum number of iterations was 10000 and the function tolerance was $10^{-6}$.

A creep experiment was conducted to identify the parameters for the model 2 (GKV). Instead of using an external linear actuator to compress the DEA, the active force of the DEA itself was used. For this purpose, a second actuator was used. The experiment assumed that a constant voltage produces a constant stress in order to collect data according to Figure 3 c). During the creep experiment, twelve measurements, at 100 V intervals, were carried out. At the beginning of each measurement, two voltage steps with a length of 30 s and a pause of 30 s were applied to the actuator in order to precondition the material and reduce the influence of the long-term viscoelastic effects and the loading history. Subsequently, a third voltage step was applied and the voltage was held for 600 s. To allow the material to relax, a break of two hours was taken between the different voltage values and before the next measurement the distance sensor was zeroed. During the measurements, the time, the target voltage signal, the high voltage applied and the deformation of the actuator were measured with a frequency of 1 kHz. Here, the measured deformation value corresponded directly to the change in height $\Delta z$. The setup described in [52] was used to record the data and deformation values of the DEA are to be considered with a resolution of 0.15 µm and an accuracy of 19.4 µm. After the experiment, the data was processed with MATLAB. At the beginning, the preconditioning steps were removed from the data, which was down sampled afterwards by a factor of 10 using the MATLAB function downsample. The data was then filtered with a moving median filter. The smoothing factor was dependent on the applied voltage value. At 100 V it was 0.04, at 200 V 0.02 and above 200 V 0.01. In addition, a delay of 20 ms, which is due to the setup for data acquisition, was removed from the data. Subsequently, the processed data was used to calculate the applied stress $\sigma_{KV_{const}}$ and the resulting strain $\varepsilon_z$ using (1) and the measured height of the elastomer stack (10.4120 mm). Using the MATLAB function fit with the Nonlinear Least Squares method and the Trust-Region algorithm, the parameters for (10) were identified. The maximum number of iterations was defined as 10000 and the function tolerance as $10^{-6}$.

In model 3 (GKM) the identified parameters form model 1 (GM) were used. For the additional damper $d_0$ the parameter was manually selected by taking the average value of the identified damper values $d_1$ to $d_3$.

For the identification of model 4 (StSp), the data from the creep experiments was used. However, the complete data set, including the preconditioning steps, was used, to provide the algorithm with as much information as possible. An additional pre-processing step was taken by using the Hookean model from Table SI, with the identified parameters from the compression test, to predict a static deformation with the down sampled input voltage. The static deformation was then also filtered with a moving median filter to ensure that the input and output had the same dynamics during identification before being combined into separate data objects using the MATLAB function iddata (for each voltage one data object). Subsequently, the resulting data objects were used to identify process models of type PZ3 using the MATLAB function procest. A PZ3 process model structure is shown (13) [53].

$$G(s) = \frac{K_p(1+s\cdot T_z)}{(1+s\cdot T_{p1})\cdot(1+s\cdot T_{p2})\cdot(1+s\cdot T_{p3})} \quad (13)$$

Then the MATLAB function isdd was used to convert the process models into state space representations and the function c2d to discretise them. The intermediate step via the process models was chosen to improve the model quality. During the identification of the process models, no disturbance model was used and the focus was set on simulation. The search method was specified as auto and the maximum number of iterations was defined as 10000 as well as the function tolerance as $10^{-6}$.

### *D. Dynamic implementation*

After the parameter identification was completed, the models were implemented. For models 1 (GM) and 3 (GKM), the systems of differential equations were transferred into Simulink block diagrams. The solver ODE23s and a variable step size were used to solve the resulting Simulink models. The output of both models was the strain $\varepsilon_z$, which was then converted with the original thickness $z_0$ and the number of layers $j$ into the change in height of the stacked DEA $\Delta z$. Both models were implemented with a fixed and a varying set of parameters. To implement the varying parameter set, the n-D Lookup Table block from Simulink was used. For model 4, the Hookean model was solved for $\lambda_z$ subsequently $z_0$ and $j$ were used again to calculate the height change $\Delta z$. With the help of the Discrete State-Space Simulink block, the dynamics were then added to this predicted steady state value. Model 4 was only implemented with a varying set of parameters, using the n-D Lookup Table block. The fixed step auto solver with a step size of 0.001 s was applied to solve the model. Unlike the other models, Simscape was used to implement model 2 (GKV). Simscape is a software that allows physical simulations to be carried out in the Simulink environment. As a result, the springs and dampers of the Kelvin-Voigt model could be implemented as physical elements and the differential equation was conducted by Simscape. To solve the physical model the solver ODE15s with a variable step size was used. The model 3 (GKM) was implemented with a fixed and a varying set of parameters.

For the varying implementation the Simscape elements varying translational spring and damper together with the n-D Lookup Table block were used. As the output of model 3 (GKM) was directly a change in the thickness only the number of layers $j$ was used to calculate the complete change in the DEA height $\Delta z$.

*E. Dynamic optimization*

To improve the quality of the models 1-3 (GM, GKV and GKM) a parameter optimization was conducted. The optimization was implemented in MATLAB using the particle swarm optimizer [54]. Particle swarm is a global population-based algorithm proposed by Kennedy and Eberhart [55] and extended by Mezura-Montes and Coello Coello [56] as well as Pedersen [57]. For optimization, the data sets of the creep identification were used. In the first step, the respective model parameters were optimized for each measured voltage value. Therefore, the implementation with a fixed set of parameters was used and the respective particle was loaded into the Simulink model via a SimulationInput object. The starting values for the optimization were always the identified parameters. During the optimization it were not the parameters themselves that were optimized but a factor by which the parameters were multiplied. The best particle from the previous optimization, except for the 100 V measurement, was used to initialize the next particle swarm. In a second step, a mean value was calculated for each parameter of the models. These averaged model parameters were then optimized again over all twelve measurements. Again, the implementation with a fixed set of parameters was used. For the optimizations, except for $d_0$ from model 3 (GKM), -40 % and +40 % were always used as the upper and lower parameter limits to avoid over fitting. Since the starting value of $d_0$ does not correspond to any identified parameter, the limits were set to $10^{-4}$ and $10^{4}$. Each optimization was run for 10000 iterations with a functional tolerance of$10^{-6}$ and 1000 stall iterations. The inertia range of the algorithm was restricted from 0.5 to 1. After an optimization was completed, an attempt was made to further reduce the function value with the hybrid function fmincon in the area of the global minimum. The mean absolute error (MAE) was used as the function value. To improve the performance of the optimizations, they were parallelized using the local MATLAB parallel pool. In addition, a timeout function was implemented, which stopped the evaluation of the current particle if it lasted longer than 10 s. If the actual simulation time of a particle did not correspond to the specified simulation time, the function value of this particle was multiplied by $10^{99}$. Thus, particles should be sorted out which produce an oscillating or generally more computationally intensive system.

*F. Dynamic validation*

In order to validate the models, rectangular, triangular and sinusoidal signals were used. Each signal type was applied with 0.1 Hz, 1 Hz as well as 10 Hz and had an offset of 650 V and an amplitude of 400 V.

Thus, all validation signals contain frequencies and amplitudes that were not included in the identification/optimization data. During the validation, both the implementations, fixed and with a variable parameter set, were used. To assess the quality of the models, the coefficient of determination $R^2$ was utilized.

## IV. RESULTS

At the beginning of this chapter, the results of the static identification will be presented. The data should be considered with an accuracy of 5 µm and 0.02 N, respectively. Then the dynamic identification is considered before the results of the validation and simulations are described, where the data of the generalized Maxwell should be also considered with an accuracy of 5 µm and 0.02 N, respectively and the data of the generalized Kelvin-Voigt, the generalized Kelvin-Maxwell, the state space model as well as the validation should be considered with an accuracy of 19.4 µm.

*A. Static identification*

The During the static identification the parameters from Table SIII in the supplementary materials were used, describing the actuator from the compression experiments. Figure 6 shows the results of the fit of the different static models from (4) and Table SI.

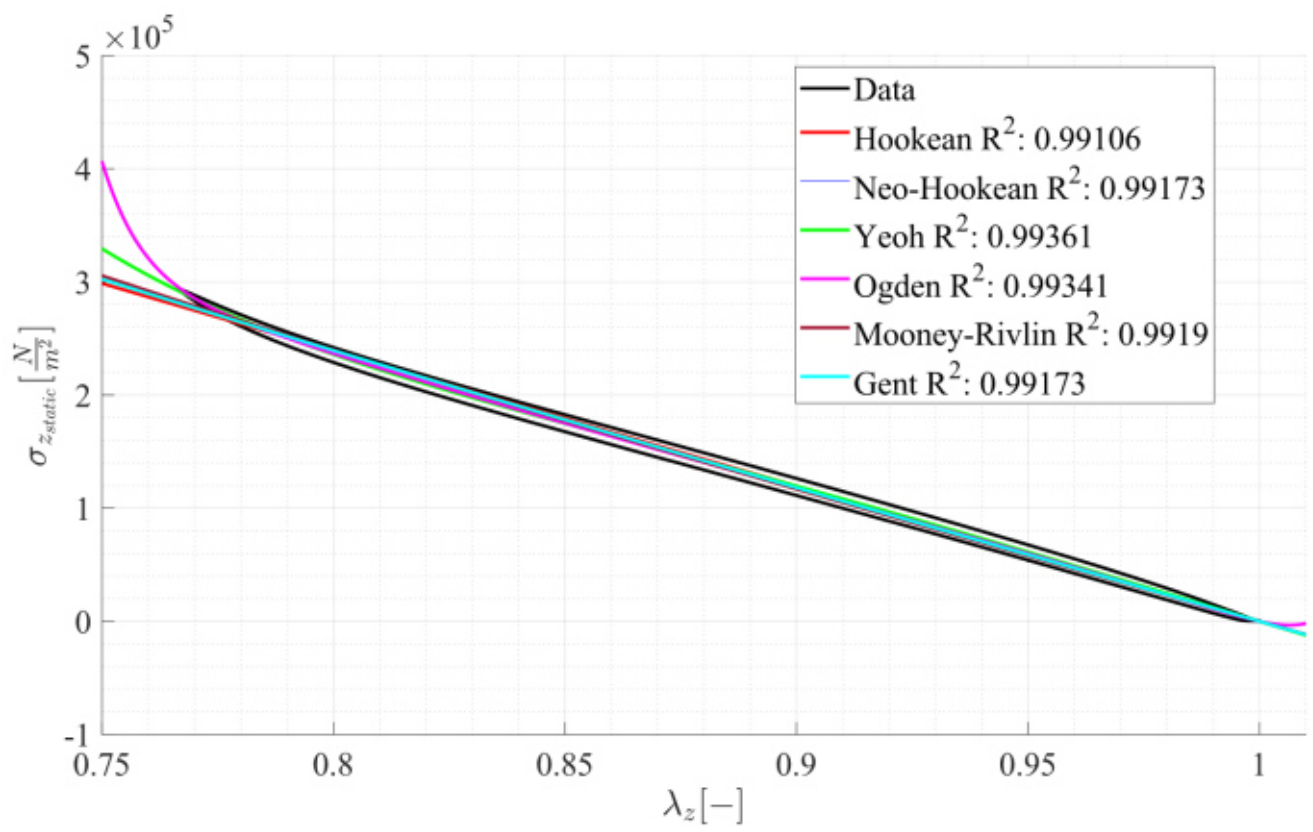

**Fig. 6.** Fit of static models

The black line represents the experimental data and a small hysteresis and a moderate non-linear behavior can be seen in Figure 6. In the beginning, all models are close to each other. Differences only arise in the end, where the Hookean, Neo-Hookean, Mooney-Rivlin and Gent model extrapolate the data almost linearly and the Yeoh and Ogden model show a clearly non-linear extrapolation. In the metrics, the models differ only from the third decimal digit onwards. With an $R^2$ of 0.99361, the Yeoh model represents the best fit. The Table SIV in the supplementary materials lists the parameters of the static models. According to the positive consideration of the compressive stress in Figure 6, the negative signs of the parameters result. All parameters are presented five significant digits.

*B. Dynamic identification*

In Figure 7, the fit of the model 1 (GM) from (12) is shown also using the parameters from Table SIII together with the fitted Young's modulus from Table SIV. Figure 8 plots the results for the fits of the model 2 (GKV) from (10) for the different voltage values. The results of the identification of the model 4 (StSp) can be seen in Figure 9. For these identifications the parameters in Table SV in the supplementary material were used, describing the actuator of the creep experiment.

**Generalized Maxwell model:** The black line Figure 7 represents the experimental data. It can be seen that after about 50 seconds most of the relaxation is completed. Figure 7 also proves that two Maxwell elements, in contrast to three elements, are not sufficient to adequately represent the course of relaxation. The identified parameters are listed in Table SVI in the supplementary materials. All parameters are presented five significant digits.

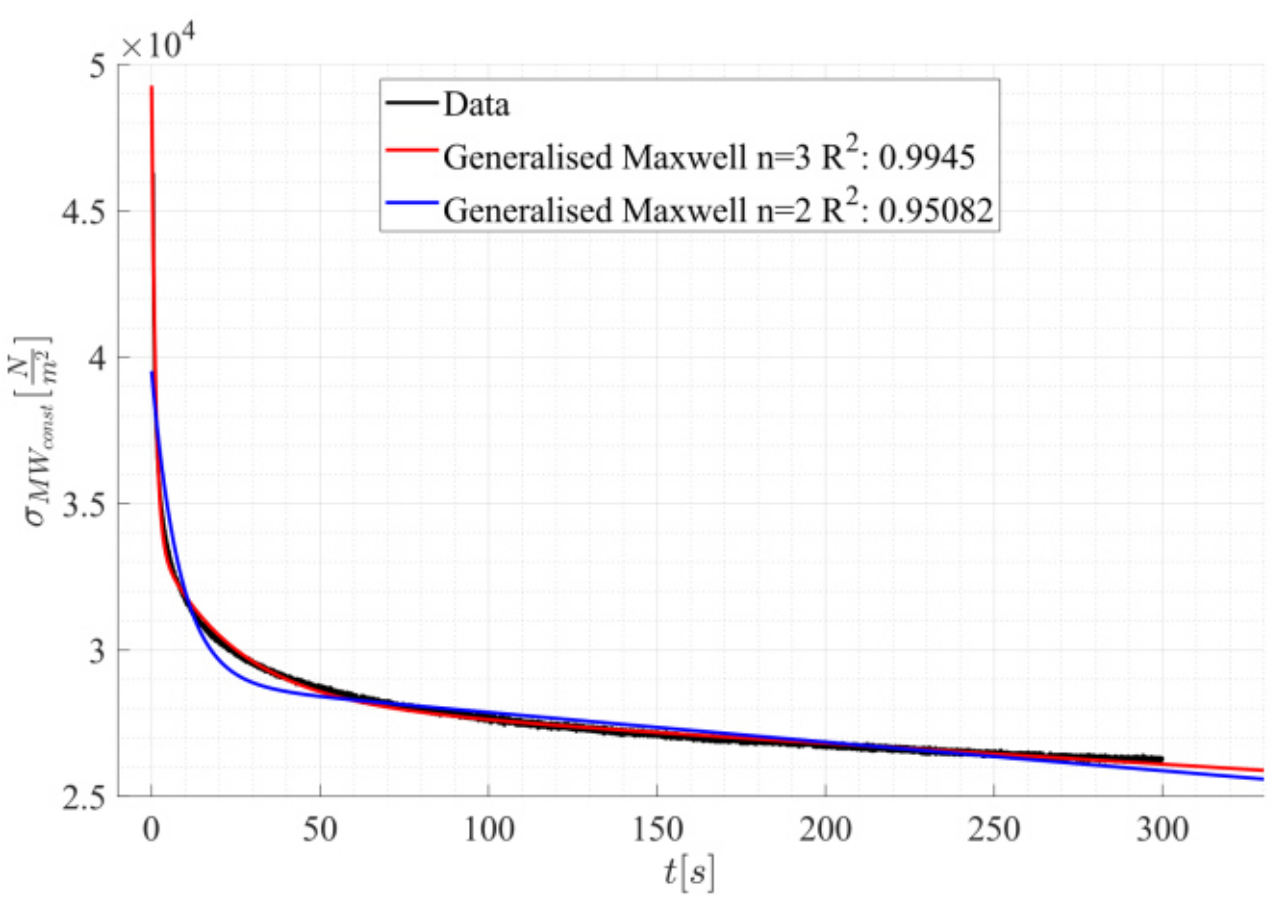

**Fig. 7.** Fit of generalized Maxwell model

**Generalized Kelvin-Voigt model:** Due to visibility Figure 8 displays the results of the model 2 (GKV) identification for 600 V and 1200 V. The full plot is available in Figure S1 in the supplementary materials.

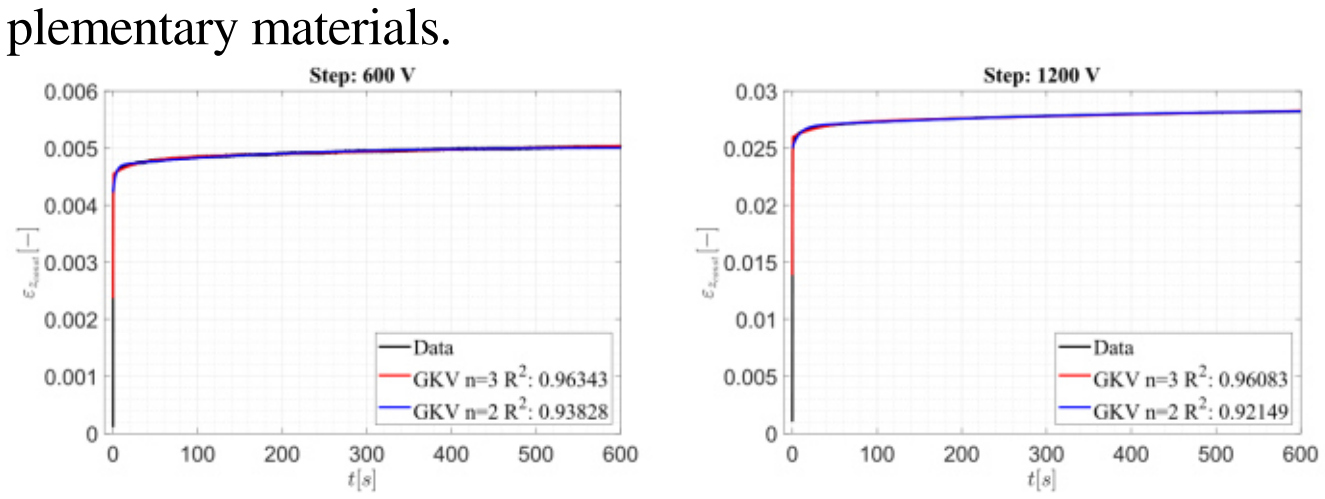

**Fig. 8.** Fits of generalized Kelvin-Voigt (GKV) model

It is visible that the $R^2$ of the fit increases with the voltage level. Only the third step, after preconditioning of the material, was used for identification. Furthermore, it can be seen that three Kelvin-Voigt elements result in a better fit then the two-element model. However, the $R^2$ value is not as high as with the Maxwell model. The Table SVII in the supplementary material shows the respective parameters per voltage value. For reasons of visibility, the confidence bounds are not listed. All parameters are presented five significant digits.

**State space model:** Similarly, for model 4 (StSp), only the results for 600 V and 1200 V are shown in Figure 9. The full plot is available in Figure S2 in the supplementary materials.

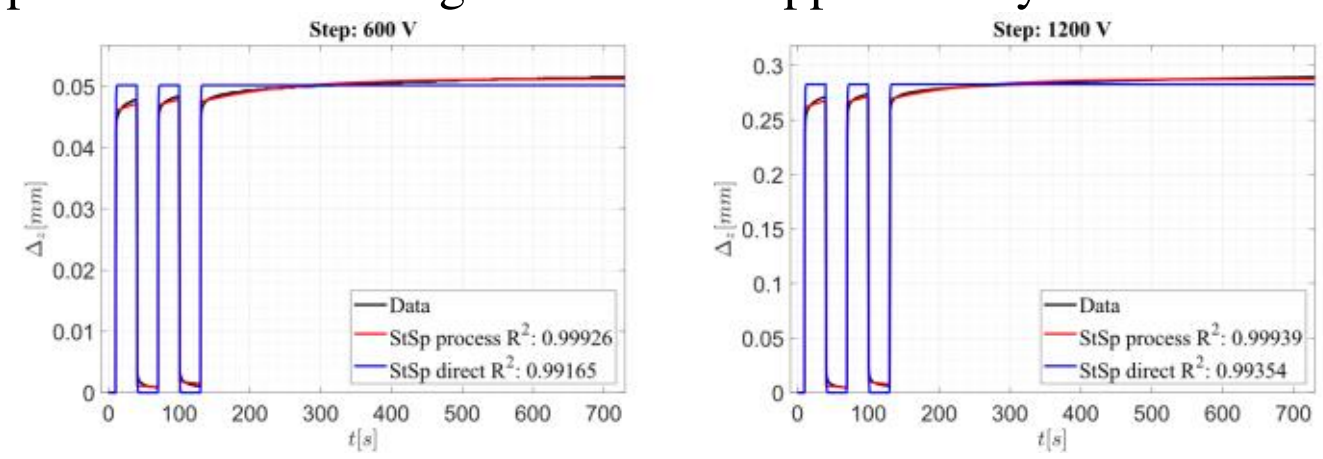

**Fig. 9.** Fits of third order discrete state space (StSp) model

In the fits of model 4 (StSp) shown in Figure 9, the model quality also increases with voltage. Furthermore, it can be seen that the state space model, which was generated from the process model in (13) (StSp process), can represent the course of the creep much more accurately than the directly identified state space model (StSp direct). The Table SVIII in the supplementary material shows the matrices of the respective state space model. $D = 0$ applies to all models. All parameters are presented with an accuracy of four digits after the decimal point.

*C. Optimized parameters*

After the identification process, the models were implemented and simulated. Once Again, the parameters were passed to the fixed parameter implementation using a SimulinkInput object. The simulation results revealed the necessity of optimizing the viscoelastic material parameters. For the optimization, the data from the creep experiments was used. In addition to the parameters of the viscoelastic models to be optimized, the parameters shown in Table SIX S5 in the supplementary materials were used for the simulation. To check the optimization results the down sampled and filtered creep data without the removed delay was used. The Figure 10 to Figure 12 show, for clarity and visibility only for 600 V and 1200 V, exemplary simulation results for the non-optimized and the optimized parameter. The full plot is available in Figure S3 to Figure S5 in the supplementary materials.

**Optimized generalized Maxwell model:** The results presented in Figure 10 show that optimization was necessary and that the model quality improved through the process. It can also be seen that the individually optimized parameters always give better or equally good results as the averaged and optimized parameter set. Even though the averaged and optimized parameter set is not as good as the individual parameter sets, it still shows an improvement compared to the original parameter set. Furthermore, the error tends to decrease with increasing voltage. The full plot is available in Figure S3 in the supplementary materials.

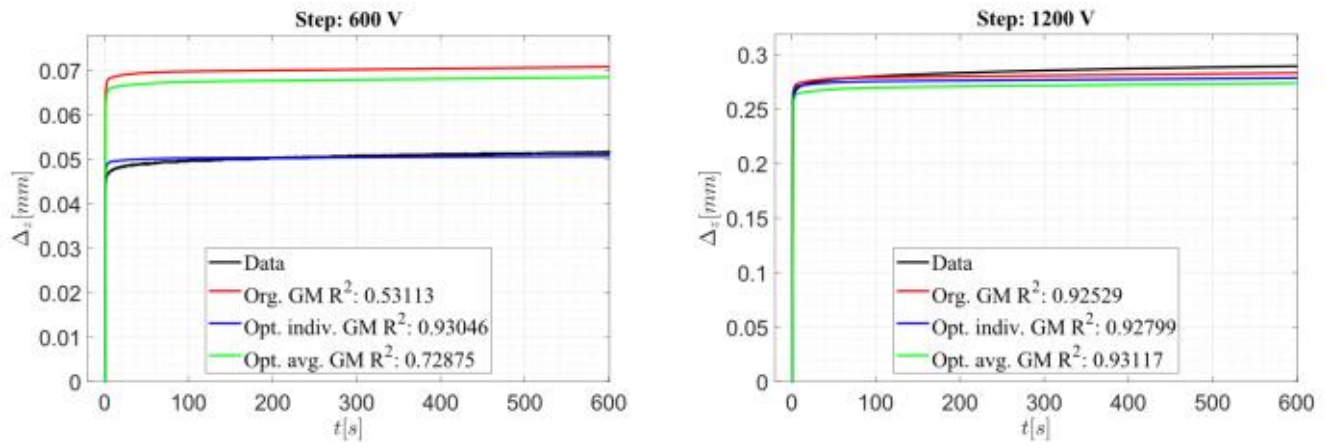

**Fig. 10.** Optimization results of generalized Maxwell (GM) model 1 with non- optimized (Org.), individual optimized (indiv.) as well as averaged and optimized parameters (avg.)

**Optimized generalized Kelvin-Voigt model**: In Figure 11, it can be clearly seen that the optimization has also improved the model quality at all voltage levels. It is worth mentioning that the averaged and optimized parameter set always gives better results than the individually optimized parameters. Furthermore, it can be seen that the originally identified parameters are further away from the data than those of the model 1 (GM) in Figure 10. Once again, the quality appears to improve with increasing voltage, particularly for the averaged and optimized parameter set. The full plot is available in Figure S4 in the supplementary materials.

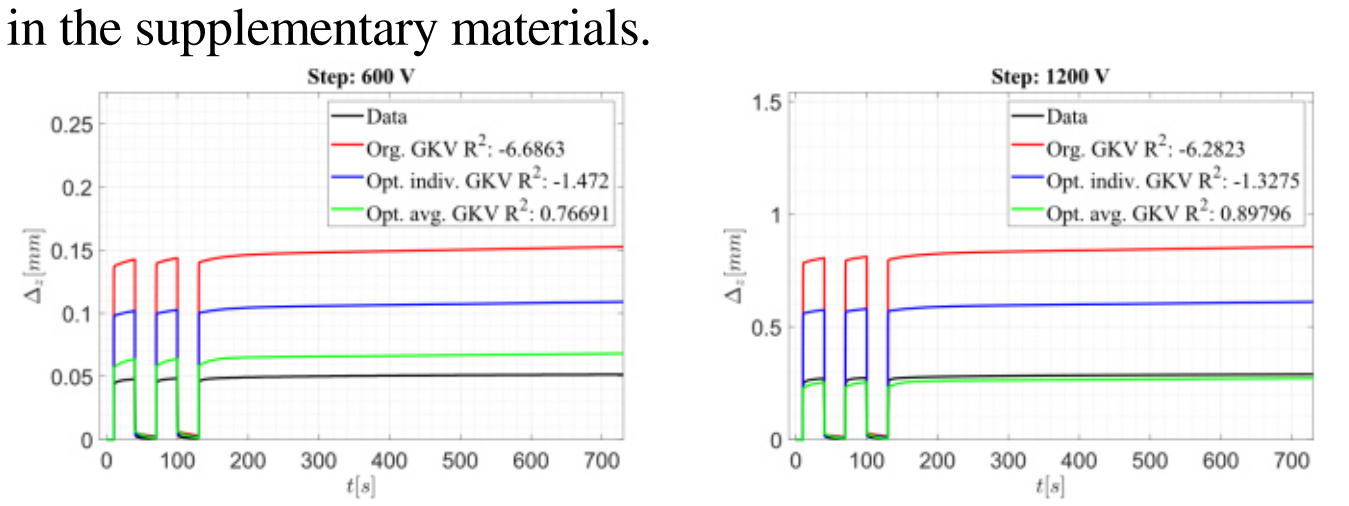

**Fig. 11.** Optimization results of generalized Kelvin-Voigt (GKV) model 2 with non- optimized (Org.), individual optimized (indiv.) as well as averaged and optimized parameters (avg.)

**Optimized generalized Kelvin-Maxwell model:** Figure 12 demonstrated that the model quality of the model 3 (GKM) has also improved as a result of the optimization. A notable aspect is the influence of the manually selected starting value for $d_0$, which can be seen in the changed dynamics of the non-optimized parameters. The individual optimized parameters deliver better or equal results for all voltages. The full plot is available in Figure S5 in the supplementary materials.

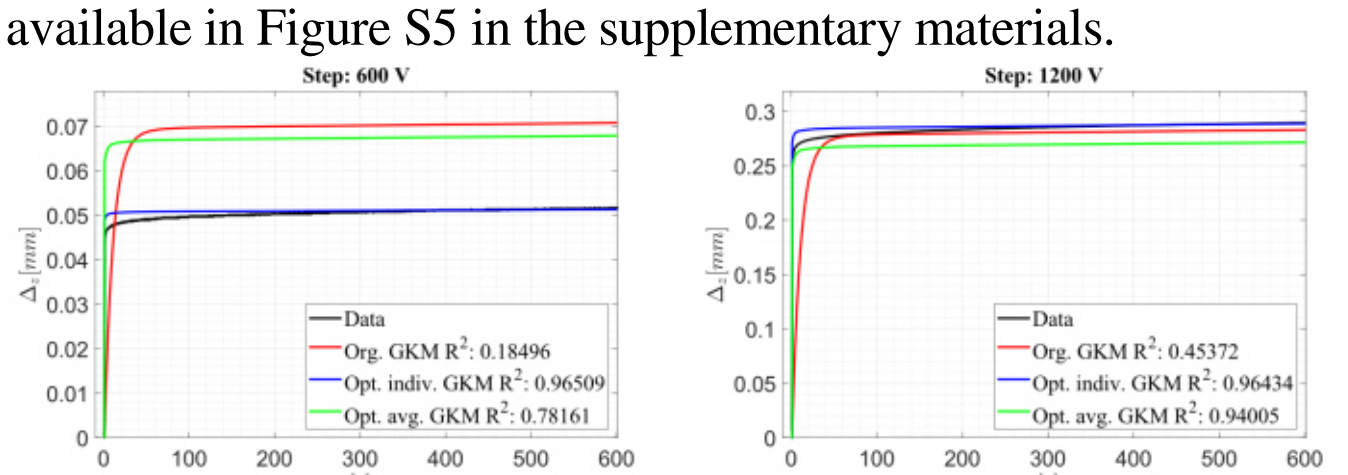

**Fig. 12.** Optimization results of generalized Kelvin-Maxwell (GKM) model 3 with non- optimized (Org.), individual optimized (indiv.) as well as averaged and optimized parameters (avg.)

*D. Validation*

As mentioned in Section 3, the models were validated with three different signal types and frequencies which were not part of the identification or optimization. The respective output signals for the simulations are shown in Figure 13 to Figure 16. The resolution of the time axis at the respective frequency is selected so that one period is visible. For the validation the actuator of the relaxation experiments was used. During the validation measurements there was an averaged temperature of 21.4 °C and an averaged humidity of 30.7 %. After each validation measurement a break of two hours was taken. For the simulation, the parameters from Table SX in supplementary materials, along with the identified or optimized parameters, were employed.

**Validation generalized Maxwell model:** Figure 13 reveals two notable aspects. Firstly, no matter what the signal shape or implementation technique, the quality of the model decreases with increasing frequency. Also, at 10 Hz it is clearly visible that the models do not represent the delay in the data acquisition. Secondly, the model gives better results for signals with continuous slope (sine and triangle) than for a signal with rapid change (rectangle). Furthermore, it becomes apparent, especially with the rectangular signal, that the model with the individually optimized parameters, which are stored in a look-up table (GM look-up), struggles to match the static final value.

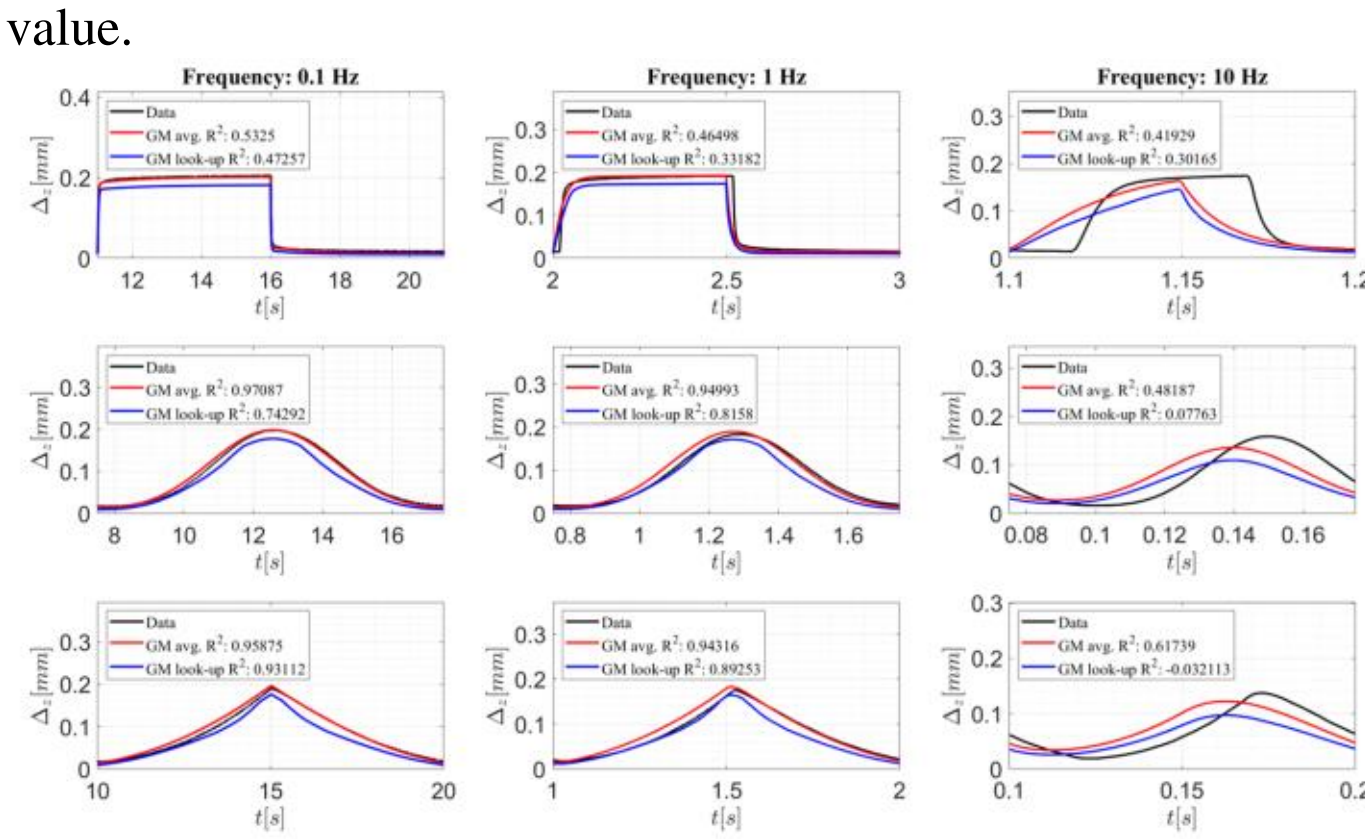

**Fig. 13.** Validation results of generalized Maxwell (GM) model 1 with fix set of parameters (avg.) and varying set of parameters (look-up) implemented

**Validation generalized Kelvin-Voigt model:** Also, the model 2 (GKV) in Figure 14 shows the phenomenon that the model quality decreases with increasing frequency and the delay is clearly visible in the rectangular 10 Hz signal. It also shows that, except for 10 Hz, the signals with a continuous slope give better results. Notable are the peaks and bumps in the course of the implementation with the look-up table (GKV look-up), which also seem to increase with frequency. Also, the distance between the two forms of implementation is much bigger in Figure 13 than in Figure 14.

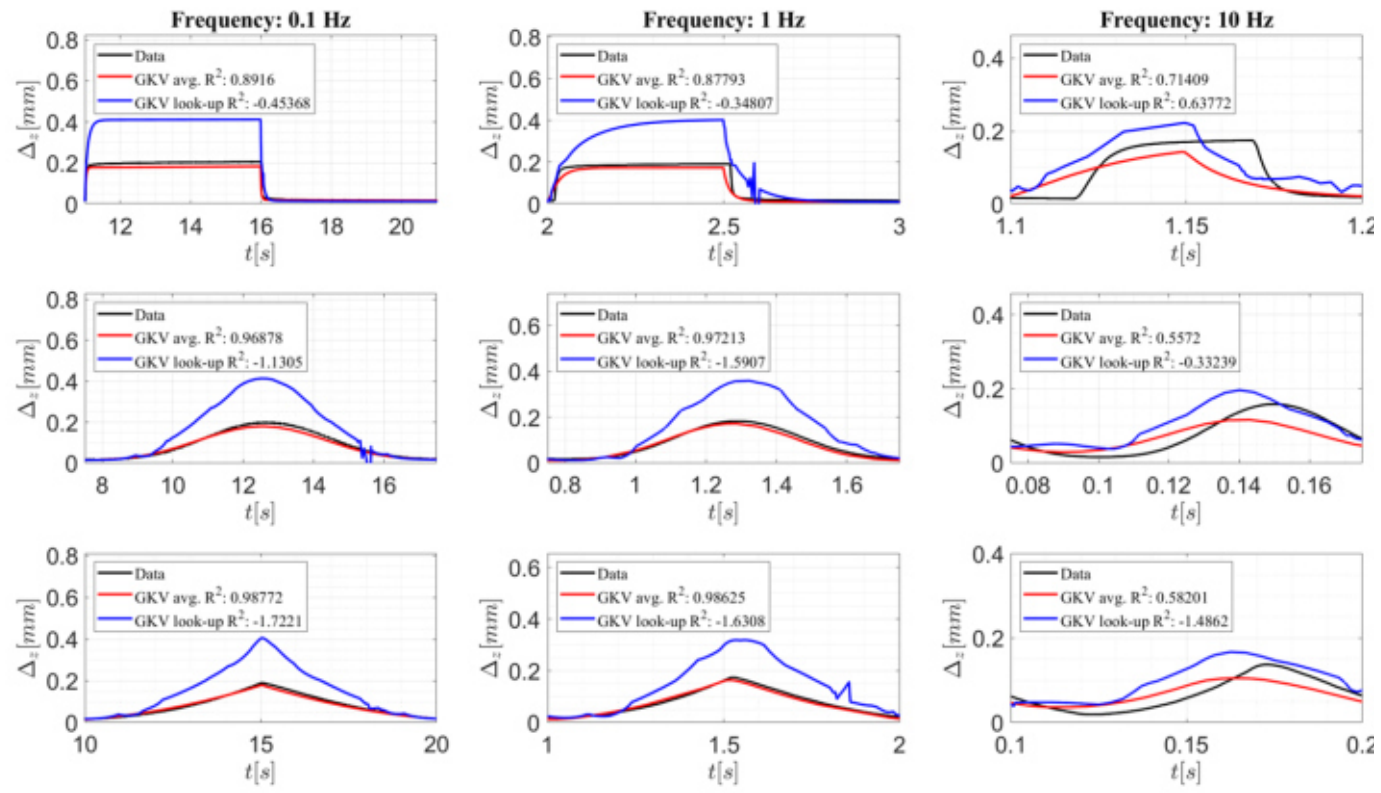

**Fig. 14.** Validation results of generalized Kelvin-Voigt (GKV) model 2 with fix set of parameters (avg.) and varying set of parameters (look-up) implemented

**Validation generalized Kelvin-Maxwell model:** For the model 3 (GKM) in Figure 15, the quality also decreases with the frequency and the acquisition delay is again visible (rectangular 10 Hz signal). Also, the signals with a continuous slope show better results again, except for 10 Hz. In contrast to Figure 13, the implementation with the look-up table shows better results. It should also be mentioned that in both Figure 13 and Figure 15, the $R^2$ value for the 0.1 Hz and 1 Hz rectangular signals does not appear to match the qualitative curve.

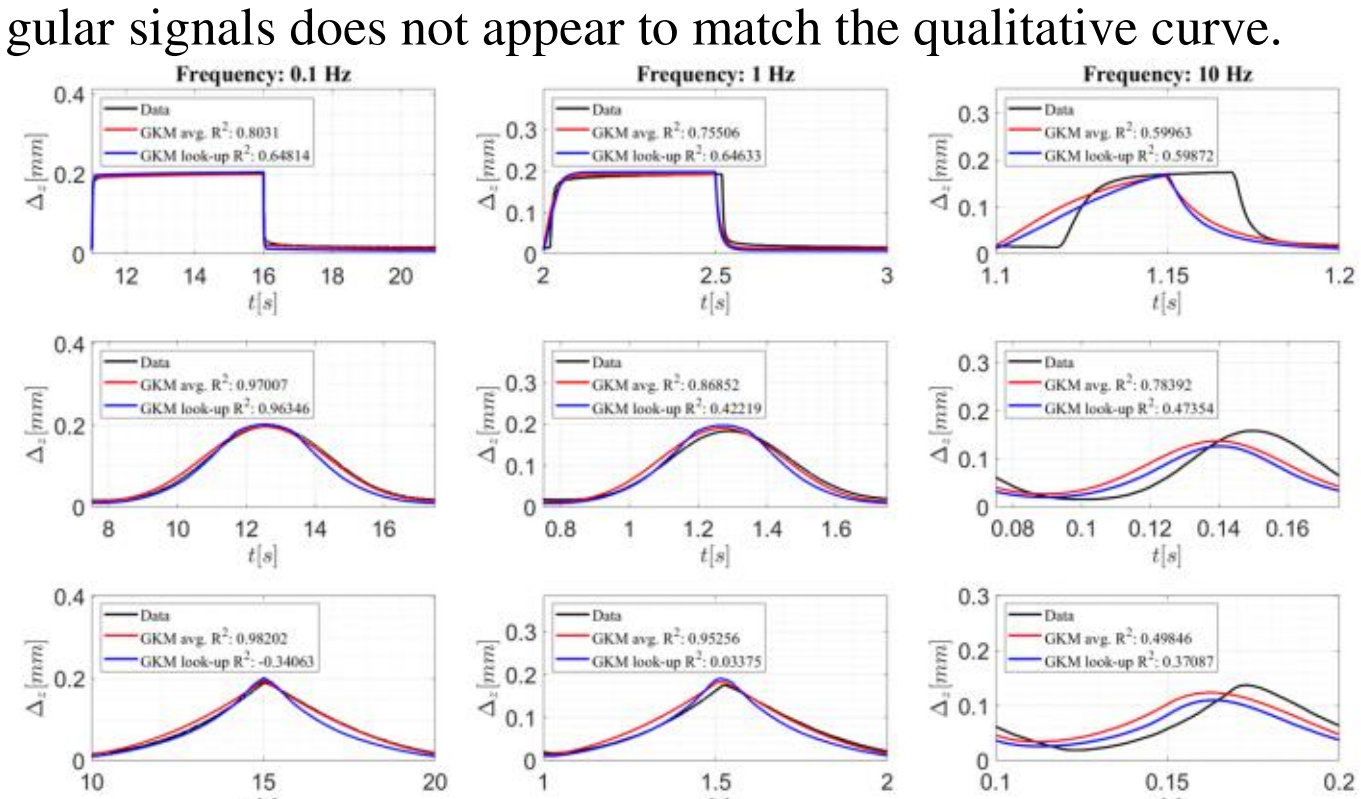

**Fig. 15.** Validation results of generalized Kelvin-Maxwell (GKM) model 3 with fixed set of parameters (avg.) and varying set of parameters (look-up) implemented

**Validation state space model:** The quality of the model 4 (StSp) in Figure 4 also decreases with increasing frequency and the acquisition delay is also present. What is interesting, however, is that there is almost no difference between the different signal types. The only thing that emerges is that the two signal types that previously delivered better results (sine and triangle) show oscillations in their model response as the frequency increases.

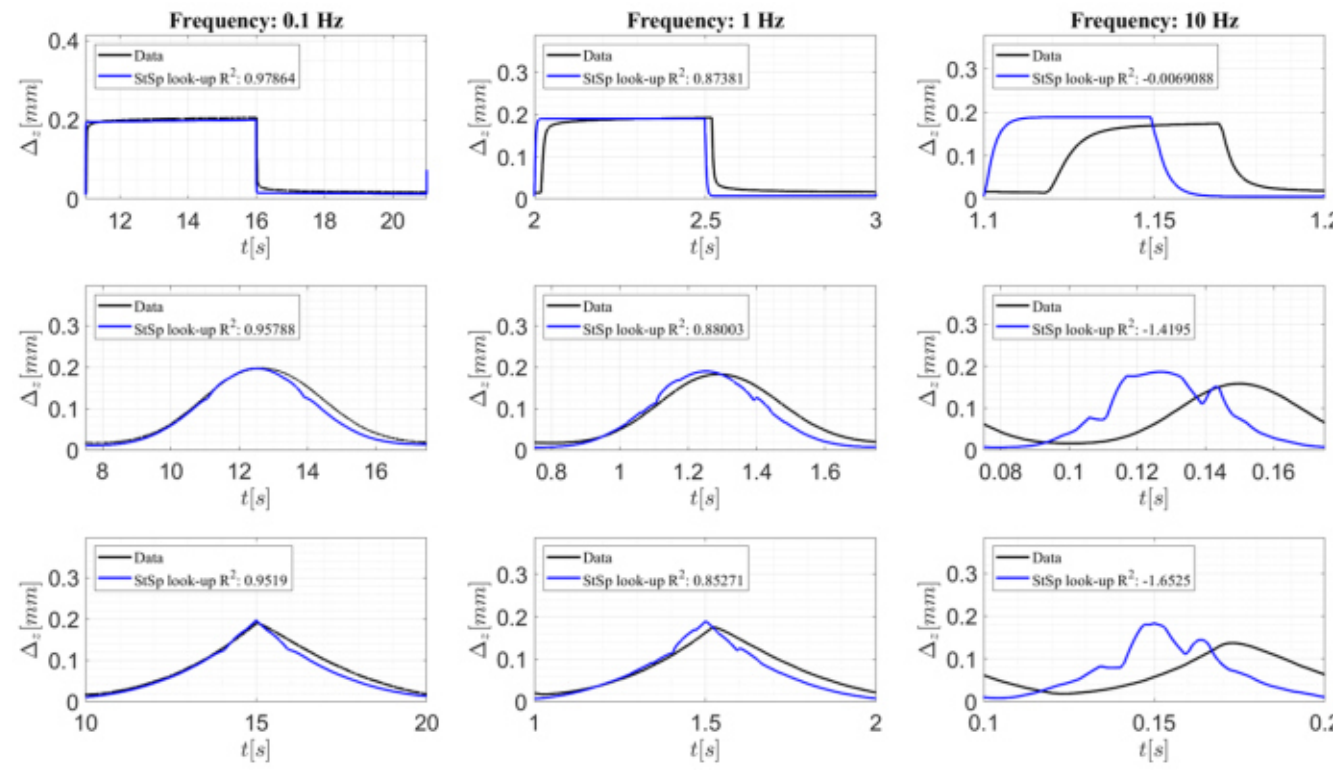

**Fig. 16.** Validation results of third order discrete state space (StSp) model 4 with varying set of parameters (look-up) implemented

## V. Discussion

The assumption of a uniaxial state and the description of the actuator by the behaviour of a single layer are, of course, simplifications, but they are supported by other works e.g. [21]. Considering the actuator as a unit during characterisation is a common assumption in a phenomenological approach. For example, in [58] a characterisation was also carried out for the complete actuator.

For the quasi-static test, a compression of 20 % was chosen to check whether the linear range of the actuator's deformation also extends beyond 10 %. Apart from that, this methodology is also supported by [21]. Figure 6 shows that the assumption of a linear stress-strain relationship is valid even up to 15 % to 20 %. Furthermore, viscoelastic effects can be seen in Figure 6, although the compression test was performed with a low deformation rate. These phenomena are consistent with literature information on the deformation of elastomers (compare Figure 3 a) and b)) [11, 26]. These results indicate that even in a smaller deformation range, viscoelastic effects should also be considered. At the end of the third cycle, a deformation of 0.0366 mm remains, which corresponds to approx. 0.4 % of the height of the elastomer stack and was calculated the same way as the initial length in Section 3. Hence the fitted models represent a trade-off between loading and unloading and approximate the true elastic deformation. The determined material parameters still have their validity and significance, since e.g. the determined elasticity modulus of $Y_{Exp} = 1.1947 \cdot 10^6$ Pa corresponds to the elasticity modulus specified by the manufacturer of $Y_{Man} = 1.4 \cdot 10^6$ Pa [33]. Thus, within the investigated deformation range, the linear elastic model represents the best compromise between complexity and model quality. However, if a better representation of the data in the range of 20 % compressions or more is required, the Yeoh model with three parameters seems to be more computationally efficient than the Ogden model with its six parameters.

The results from Figure 7 show an improvement of the fit of approximately 5 % by adding another Maxwell element. However, when using four Maxwell elements, the $R^2$ value decreases to 0.5083. Thus, the generalised Maxwell model with three elements represents the best structure.

This finding is consistent with [59], where also three elements were used to represent a frequency range up to 100 Hz for a silicone-based actuator. Since the fitted Young's modulus from the compression tests was used for the preparation of the measured data, any uncertainties may also be present in the parameters of the Maxwell model. However, the applied procedure is supported by [21].

Compared to the model 1 (GM), the fits of the model 2 (GKV) in Figure 8 are slightly less accurate. There are two possible reasons for this. On the one hand, the measurement data used for the fitting exhibits more noise. The signal has a fluctuation of 4.7 μm [52]. On the other hand, the GKV model in Equation 10 assumes that the induced stress is constant. In this creep experiment, however, the DEA is electrically activated and the force $F_{active}$ changes with the deformation. This variation in force is due to the reduction in the distance between the electrodes and the increase in electrode area. Nevertheless, it is also shown here that three elements represent the best model structure. The use of a second actuator for the creep experiments had setup-related reasons. The second actuator was permanently mounted in its test bench [52] to avoid errors arising from different actuator positions under the laser sensor during various measurements. An effort was made to keep the influence on the identification, optimization, and subsequent validation processes as low as possible by using a second actuator of the same type and taking into account the slightly different dimensions between the two actuators (compare Table SIII, Table SV, Table SIX and Table SX in the supplementary materials).

With regard to the $R^2$ value model 4 (StSp) in Figure 9 represents the best result during the identification. These good results of the identification process have also made it unnecessary to further optimise the model. The difference between the directly identified and the process model generated state space representation is most likely due to the numeric method used during the identification process. The conversion from a process model to a state space representation was carried out because it is a very common model structure that allows investigations into stability etc. and is well suited for subsequent research, e.g. for the design of a controller [60]. In this study, it was deliberately avoided to identify an averaged or generally valid set of parameters, since the parameters do not correspond to any real physical quantity and an identification process of a process model of the type PZ3, over all voltage jumps, delivered poor results. The use of a state space representation has already been proposed by [58]. In this case, a generalised Maxwell model with n=3 was converted into a state space representation. However, the advantages of the black box identification seems to get lost, as the basic model structure does not change in comparison to a generalised Maxwell model and no additional degrees of freedom are created.

In Figure 10 to Figure 12, the red lines (org.) clearly demonstrate the necessity for an optimization process. The initial parameters of the model 2 (GKV) show the greatest deviation from the measured data.

This is due to the two points mentioned above: noise and the assumption of a constant force. To address this issue, it would be beneficial to conduct passive creep experiments as well. This could be done using calibration weights to apply a constant force similar to [61]. The most reliable set-up, of course, would be a test bench with force control. However, a passive creep experiment with the required force control is more difficult to implement. Furthermore, the optimisation was able to compensate for the deviation. The fact that the implementation with the averaged parameter set (Opt. avg. GKV) is probably due to the fact that the parameters went through two optimisations, once before and once after they were averaged. For the model 1 (GM) and model 3 (GKM) the deviations are almost identical, as they only differ due to the additional value for $d_0$. In both cases, as expected, the individually optimised parameters deliver better results. The creep data including the delay were used to check the optimisation results to find out if the stability of the models is affected by the delay [62]. No stability issues were found for the analysed step function. For all three viscoelastic models, it can be seen that the averaged and optimised parameter set (avg.) is particularly suitable for representing larger deformations.

The previous statement is confirmed by the validation of the model 1 (GM) in Figure 13. Since all three types of signals with their maximum voltage of 1050 V cause a high deformation, the averaged parameter set (avg.) can represent the data with high accuracy. The steady state error of the look-up table implementation is due to the fact that there is no breakpoint in the table for this maximum voltage value and linear interpolation is performed between two breakpoints. The interpolated parameters do not fully represent the system behaviour. By optimising the parameters for more voltage values, this steady state error could be reduced. However, this leads to a considerable expenditure of time, since on the one hand a relevant pause has to be inserted between the individual measurements for the optimisation and on the other hand the generalised Maxwell model is the most computationally intensive model. Table II shows the time taken by the four different models to simulate all validation signals, with a total simulation time of 153.75 seconds using a PC with an Intel Core i7-10510U CPU with 2.30 GHz and 16 GB RAM. The values are presented five significant digits.

TABLE II
COMPUTATION TIME OF THE MODELS

| | Model 1 (GM) | Model 2 (GKV) | Model 3 (GKM) | Model 4 (StSp) |
|---|---|---|---|---|
| Computation time | 30.896 min | 11.300 sec | 13.608 min | 0.6233 sec |

Figure 13 also confirms that the acquisition delay does not impact the stability of the model 1 (GM). The delay primarily leads to a decrease in the model's accuracy at higher frequencies, as according to [59], a GM model with three elements should be able to reproduce frequencies up to 100 Hz.

Another explanation could be that the model does not consider the electrical characteristics of the DEA or the test bench. Due to the presence of a current limiting resistor [52] and its feeding line, electrode and connection resistances (combined to a series resistance), the DEA itself is a low-pass filter whose cut-off frequency is around 40 Hz, but may be even lower as the series resistance of the DEA is not known. The fact that the sine and triangle profiles give better results is probably due to the fact that the viscoelastic effects are reduced by the slower deformation speed due to the continuous slope [25].

The validation of the model 2 (GKV) in Figure 14 reaffirms that the averaged parameter set is well suited to describe higher deformations. The stability problems of the look-up table implementation could be due to the delay. However, this is contradicted by the fact that the system is stable for rectangle and triangle waveforms at 0.1 Hz and for the averaged parameter set at all frequencies and waveforms. Therefore, it is plausible that the stability problems are due to the fact that the parameters in the look-up tables do not adequately describe the system, as already seen in Figure 11, and that the resulting linear interpolated parameters do not fit at all. This phenomenon is then amplified by the variation of the non-matching parameters by the look-up tables depending on the input voltage. The variation speed further effects negatively, which explains the poorer results at higher frequencies. The discrepancy between the $R^2$ values and the qualitative trajectories of the 10 Hz signals in Figure 14, also compared to Figure 13, is due to the sensitivity of this metric to extreme outliers [63]. The signal forms sine and triangle contain due to their generation by the Sine Wave and the Triangle Generator block more periods than the rectangular form, which was generated manually by the Signal Generator block. This leads to more peaks in the error value and a worse $R^2$ value.

The validation results of the model 3 (GKM) in Figure 15 are similar to those in Figure 13, which is expected since these two models differ only in one additional parameter. However, this one additional degree of freedom leads to a reduction of the steady-state error of the look-up table implementation for the rectangular profile. Furthermore, the additional damper $d_0$ leads to a reduction in the stiffness of the system, which in turn approximately halves the computation time, as can be seen in Table II. The poorer $R^2$ values of the continuous slopes at 10 Hz as well as the discrepancy between the $R^2$ values and the qualitative course at rectangular 0.1 and 1 Hz signal are again due to the characteristics of the metric described above.

Regarding the validation of model 4 (StSp) in Figure 16, the instabilities observed at 10 Hz are also due to the look-up table implementation and the resulting rapid variation and combination of linearly interpolated parameters. The reason why the oscillations do not occur with the rectangular signal is probably that its shape is closest to the identification data.

It should be mentioned that in Figure 16 only nine instead of 18 simulations had to be carried out for the state space representation. For a better comparison, the time from Table II should be doubled to 1.2465 s.

## VI. Conclusion

In this study, various static and dynamic models and implementation methods for the simulation of commercially available silicon-based stacked DEAs were investigated. The objective was to find out how these actuators can be modelled in the most effective way. For static modelling, it can be clearly stated that Hookean law is completely sufficient for the deformations of such an actuator and hyperelastic models should only be considered if the deformation range is larger than 20 % compression. In this case, the Yeoh model with three cores represents the best compromise between hyperelasticity and low complexity. For the dynamic modelling, it has turned out that the choice of modelling and implementation method strongly depends on the later purpose of the model. If the final objective is to design a controller based on the developed model, a black box identification with a process model and a conversion into a state space representation is to be preferred. This approach has the advantage of being resource-friendly and easy to apply. Furthermore, it is possible to switch easily between a continuous and a discrete representation, e.g. if a real-time simulation needs to be performed. It also delivers very good results for all waveforms in the low frequency range, which correspond to the frequency range of other proposed models [18, 19, 25]. Furthermore, the model response can be improved at higher frequencies or other waveforms if they are included in the identification data.

If the final objective is to study the properties of the actuator, the viscoelastic models are preferred. It has been demonstrated that for the identification of the passive dynamic material behaviour, passive experiments should be carried out in any case in order to obtain a good starting parameter set. Furthermore, it was shown that by optimizing the parameters the model quality can be improved and that an implementation with a parameter set is significantly more stable. An averaged and optimized parameter set is particularly suitable for representing higher deformations. In addition, it is also suitable for simulating frequencies up to 10 Hz, whereby the integration of a possible acquisition delay and the electrical characteristics should be considered what will be part of future work. With their very good results in the frequency range up to 1 Hz, the three investigated models are also in the range of the other referenced models. Comparing the three viscoelastic models with respect to computational time and simplicity of identification process, the generalized Kelvin-Maxwell model with three Maxwell elements represents the most favourable model structure. In summary, it can be stated that for the development of soft robots on the basis of DEAs or, in general, an application with artificial muscles, methods of system identification are completely sufficient and more effective.

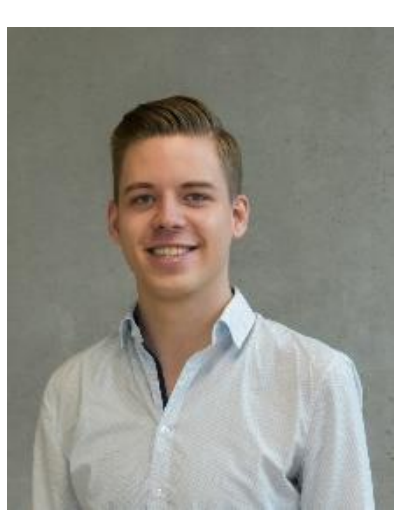

**Lukas Sohlbach** received the B. Eng. degree in interdisciplinary engineering contemporary topic mechatronics from the RheinMain University of Applied Sciences, Rüsselsheim, Germany in 2017, the M. Sc. degree in mechatronics and robotics from the Frankfurt University of Applied Sciences, Frankfurt/Main, Germany in 2019.

He is currently conducting his Ph.D. cooperatively between the Frankfurt University of Applied Sciences and the University of Cadiz, Cadiz, Spain. His research focuses on the control of soft robots and the design of mobile robot platforms.

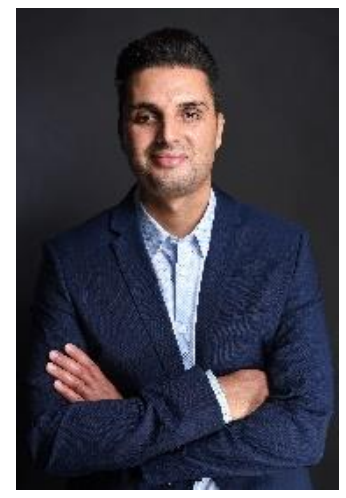

**Hamza Hobbani** received the B. Eng. degree in mechatronics in 2016 and the M. Sc. degree in mechatronics and robotics from the Frankfurt University of Applied Sciences, Frankfurt/Main, Germany in 2018.

He is currently working as a software developer at Robert Bosch GmbH conducting his Ph.D. cooperatively between the University of Cadiz, Cadiz, Spain and the Frankfurt University of Applied Sciences. His research focuses on autonomous navigation of mobile robot platforms.

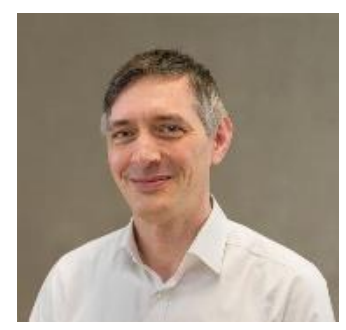

**Christopher Blase** received his Ph.D. degree in cell biology in 2009 from Goethe University, Frankfurt/Main, Germany. During his Postdoctoral position at the Institut für Materialwissenschaften, Frankfurt University of Applied Sciences, Frankfurt, Germany he worked on characterizing the mechanical properties of biological tissues from ultrasound imaging.

He is currently a Senior Scientist with the Personalized Biomedical Engineering Lab, Frankfurt University of Applied Sciences, Frankfurt/Main, Germany. His research focuses on the determination of mechanical properties of human tissues from in vivo measurements.

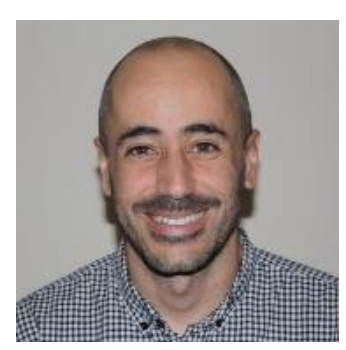

**Fernando Perez-Peña** received the Engineering degree in Telecommunications from the University of Seville, Seville, Spain and his Ph.D. degree (specialized in neuromorphic motor control) from the University of Cadiz, Cadiz, Spain in 2009 and 2014 respectively. In 2015 he was a postdoc at CITEC, Bielefeld University, Bielefeld, Germany. He has been an Associate Professor in the Architecture and Technology of Computers Department of the University of Cadiz since 2019.

His research interests include neuromorphic engineering, CPG, motor control and neurorobotics.

**Karsten Schmidt** received his Ph.D. degree in in nuclear physics in 1995 from Goethe University, Frankfurt/Main, Germany. During his Postdoctoral position he worked at GSI, Darmstadt and the University of Edinburgh. After that we was employed at Continental Automotive Systems, Lufthansa Systems and SAP. From 2005 to 2009 we were a Professor for Automotive Mechatronics at the Coburg University of Applied Sciences.

He is currently a Professor for Mechatronics at Frankfurt University of Applied Sciences. His research interests include control systems, embedded systems and mobile robotics.

## APPENDIX A

TABLE SI

HYPERELASTIC MODELS WITH S FOR STATISTICAL OR MICROMECHANICAL MODELS, P FOR PHENOMENOLOGICAL MODELS AND H FOR HYBRID MODELS

| Model | Strain Energy | True stress | Type | Reference |
|---|---|---|---|---|
| Neo-Hookean | $W_s = \frac{\mu}{2} \cdot \left(2 \cdot \frac{1}{\lambda_z} + \lambda_z^2 - 3\right)$ | $\sigma_{z_{static}} = 2 \cdot C_1 \cdot \left(\lambda_z^2 - \frac{1}{\lambda_z}\right)$ <br> $with\ C_1 = \frac{\mu}{2}$ | S | [21, 35, 42] |
| Yeoh for N=3 | $W_s = \sum_{i=1}^{3} C_i \cdot (I_1 - 3)^i$ <br> $with\ I_1 = 2 \cdot \frac{1}{\lambda_z} + \lambda_z^2$ | $\sigma_{z_{static}} = 2 \cdot \left(\lambda_z^2 - \frac{1}{\lambda_z}\right)$ $\cdot [C_1 + 2 \cdot C_2 \cdot (I_1 - 3)$ $+3 \cdot C_3 \cdot (I_1 - 3)^2]$ <br> $with\ I_1 = 2 \cdot \frac{1}{\lambda_z} + \lambda_z^2$ | P | [15, 21, 36, 42] |
| Ogden for N=3 | $W_s = \sum_{p=1}^{3} \frac{\mu_p}{\alpha_p} \cdot (2 \cdot \lambda^{-\frac{1}{2} \cdot \alpha_p}$ $+\lambda_z^{\alpha_p} - 3)$ | $\sigma_{z_{satic}} = \mu_1 \cdot \left(\lambda_z^{\alpha_1} - \frac{1}{\sqrt{\lambda_z^{\alpha_1}}}\right)$ <br> $+\mu_2 \cdot \left(\lambda_z^{\alpha_2} - \frac{1}{\sqrt{\lambda_z^{\alpha_2}}}\right)$ <br> $+\mu_3 \cdot \left(\lambda_z^{\alpha_3} - \frac{1}{\sqrt{\lambda_z^{\alpha_3}}}\right)$ | P | [38, 42] |
| Mooney-Rivlin | $W_s = C_1 \cdot (I_1 - 3)$ $+C_2 \cdot (I_2 - 3)$ <br> $with\ I_1 = 2 \cdot \frac{1}{\lambda_z} + \lambda_z^2$ <br> $and\ I_2 = \frac{1}{\lambda_z^2} + 2 \cdot \lambda_z$ | $\sigma_{z_{static}} = C_1 \cdot \left(2 \cdot \lambda_z^2 - \frac{2}{\lambda_z}\right)$ $-C_2 \cdot (\frac{2}{\lambda_z^2} - 2 \cdot \lambda_z)$ | P | [15, 35, 39–41] |
| Gent | $W_s = -\frac{\mu \cdot J_{\lim}}{2}$ $\cdot \log(1 - \frac{1}{J_{lim}} \cdot (I_1 - 3))$ <br> $with\ I_1 = 2 \cdot \frac{1}{\lambda_z} + \lambda_z^2$ | $\sigma_{z_{static}} = \frac{J_{lim} \cdot \mu \cdot \left(\lambda_z^2 - \frac{1}{\lambda_z}\right)}{J_{lim} - I_1 + 3}$ <br> $with\ I_1 = 2 \cdot \frac{1}{\lambda_z} + \lambda_z^2$ | H | [37, 42] |

TABLE SII

DYNAMIC MODELS WITH PASSIVE FORCE OR DYNAMIC EQUATION

| No. | Static description | Dynamic description | Passive force or dynamic equation | Reference |
| --- | --- | --- | --- | --- |
| 1 | Hookean law | Generalized Maxwell with n=3 | $F_{passive}(t) = A_c \cdot \left( Y \cdot \varepsilon_z(t) + \sum_{i=1}^{3} \sigma_{MW}(t)_i \right)$ | [21] |
| 2 | Hookean law | Generalized Kelvin-Voigt with n=3 | $F_{passive}(t) = A_c \cdot \sigma_z(t)$ <br> $with\ \frac{\mathrm{d}^3\sigma_z(t)}{\mathrm{d}t^3} = \frac{1}{D} \cdot (-C \cdot \frac{\mathrm{d}^2\sigma_z(t)}{\mathrm{d}t^2}$ $-B \cdot \frac{\mathrm{d}\sigma_z(t)}{\mathrm{d}t} - A \cdot \sigma(t) + d \cdot \frac{\mathrm{d}^3\varepsilon_z(t)}{\mathrm{d}t^3}$ $+c \cdot \frac{\mathrm{d}^2\varepsilon_z(t)}{\mathrm{d}t^2} + b \cdot \frac{\mathrm{d}\varepsilon_z(t)}{\mathrm{d}t} + a \cdot \varepsilon_z(t)),$ <br> $D = d_1 \cdot d_2 \cdot d_3,$ <br> $C = k_0 \cdot (d_1 \cdot d_2 + d_1 \cdot d_3 + d_2 \cdot d_3) + k_1 \cdot d_2 \cdot d_3 + k_2 \cdot d_1 \cdot d_3 + k_3 \cdot d_1 \cdot d_{2,}$ <br> $B = k_0 \cdot (k_1 \cdot d_2 + k_2 \cdot d_1 + k_1 \cdot d_3 + k_3 \cdot d_1 + k_2 \cdot d_3 + k_3 \cdot d_2) + k_1 \cdot k_2 \cdot d_3 + k_1 \cdot k_3 \cdot d_2 + k_1 \cdot k_3 \cdot d_1,$ <br> $A = k_0 \cdot k_1 \cdot k_2 \cdot k_3 \cdot \left( \frac{1}{k_0} + \frac{1}{k_1} + \frac{1}{k_2} + \frac{1}{k_3} \right),$ <br> $d = k_0 \cdot d_1 \cdot d_2 \cdot d_n,$ <br> $c = k_0 \cdot (k_1 \cdot d_2 \cdot d_3 + k_2 \cdot d_1 \cdot d_3 + k_3 \cdot d_1 \cdot d_2),$ <br> $b = k_0 \cdot (k_1 \cdot k_2 \cdot d_3 + k_1 \cdot k_3 \cdot d_2 + k_2 \cdot k_3 \cdot d_1)$ <br> $and\ a = k_0 \cdot k_1 \cdot k_2 \cdot k_3$ | [49] |
| 3 | Hookean law | Generalized Kelvin-Maxwell with n=3 | $F_{passive}(t) = A_c \cdot ( Y \cdot \varepsilon_z(t) + d_0 \cdot \frac{\mathrm{d}\varepsilon_z(t)}{\mathrm{d}t} + \sum_{i=1}^{3} \sigma_{MW}(t)_i)$ | [20] |
| 4 | Hookean law | 3rd order state space without disturbance | $x(t + Ts) = A \cdot x(t) + B \cdot u(t)$ <br> $y(t) = C \cdot x(t) + D \cdot u(t)$ <br> $with\ Ts: sample\ time,$ <br> $A: 3x3\ matrix,$ <br> $B: 3x1\ matrix,$ <br> $C: 1x3\ matrix$ <br> $and\ D: 1x1\ matrix$ | [50] |

## APPENDIX B

TABLE SIII

ACTUATOR PARAMETERS STATIC IDENTIFICATION

| $A_c\ [m^2]$ | $L_{elast}[m]$ |
|---|---|
| $1.1387 \cdot 10^{-4}$ | $0.010412$ |

TABLE SIV

IDENTIFIED PARAMETERS OF STATIC MODELS

| Model | Parameters |
|---|---|
| Hookean | $Y = -1.1947 \cdot 10^6\ Pa$ |
| Neo-Hookean | $C_1 = -1.9622 \cdot 10^5\ Pa$ |
| Yeoh | $C_1 = -2.0653 \cdot 10^5\ Pa;\ C_2 = 1.4537 \cdot 10^5\ Pa;$<br>$C_2 = -4.6806 \cdot 10^5\ Pa$ |
| Ogden | $\alpha_1 = 153.0931;\ \alpha_2 = 5.9611;\ \alpha_3 = -76.5465;$<br>$\mu_1 = 3.4940 \cdot 10^3\ Pa;\ \mu_2 = -1.4088 \cdot 10^5\ Pa;$<br>$\mu_3 = 3.4940 \cdot 10^3\ Pa$ |
| Mooney-Rivilin | $C_1 = -1.7321 \cdot 10^5\ Pa;\ C_2 = -1.8720 \cdot 10^4\ Pa$ |
| Gent | $J_{lim} = -6.2127 \cdot 10^6;\ \mu = -3.9243 \cdot 10^5\ Pa$ |

TABLE SV

ACTUATOR PARAMETERS DYNAMIC IDENTIFICATION

| $z_0\ [m]$ | $\varepsilon_0 \left[\frac{s^4 \cdot A^2}{kg \cdot m^3}\right]$ | $\varepsilon_r\ [-]$ | $L_{elast}[m]$ | $Y[Pa]$ |
|---|---|---|---|---|
| $25 \cdot 10^{-6}$ | $3.8542 \cdot 10^{-12}$ | $2.8$ | $0.010261$ | $1.1947 \cdot 10^6$ |

TABLE SVI

IDENTIFIED PARAMETERS OF GENERALIZED MAXWELL MODEL WITH N=3

| Model | Parameters |
|---|---|
| Generalized Maxwell n=3<br>with $Y = 1.1947 \cdot 10^6\ Pa$ | $k_1 = 1.5714 \cdot 10^5\ Pa;\ k_2 = 3.0800 \cdot 10^4\ Pa;$<br>$k_3 = 8.5729 \cdot 10^4\ Pa$<br>$d_1 = 5.8299 \cdot 10^8\ \frac{Pa}{s};\ d_2 = 7.2180 \cdot 10^5\ \frac{Pa}{s};$<br>$d_3 = 9.8021 \cdot 10^4 \frac{Pa}{s}$ |

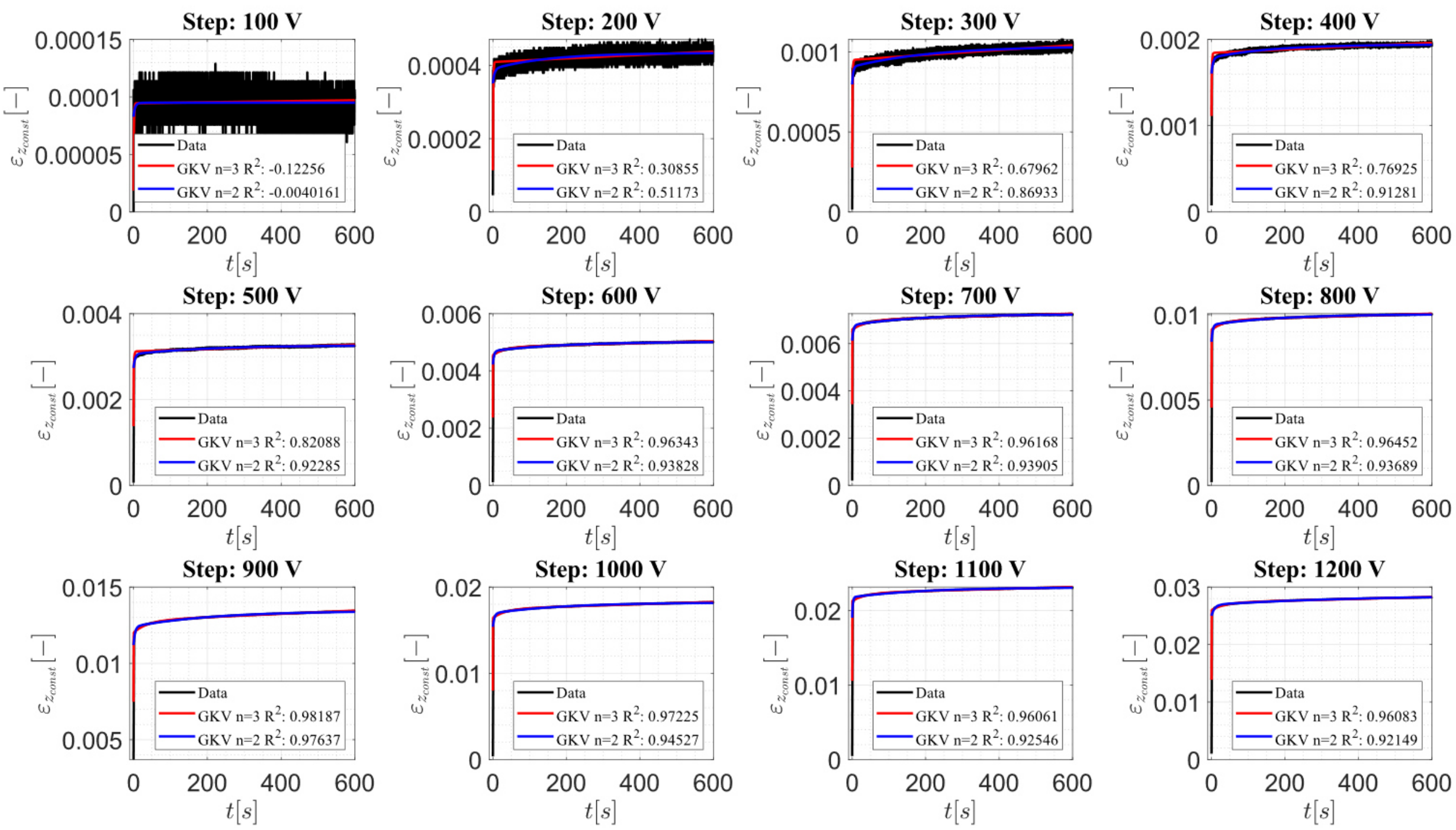

**Fig. S1.** Fits of generalized Kelvin-Voigt (GKV) model for all voltage levels

TABLE SVII

IDENTIFIED PARAMETERS OF GENERALIZED KELVIN-VOIGT MODEL WITH N=3

| Voltage [V] | $k_0\,[Pa]$ | $k_1[Pa]$ | $k_2[Pa]$ | $k_3[Pa]$ | $\alpha_1\left[\frac{Pa}{s}\right]$ | $\alpha_2\left[\frac{Pa}{s}\right]$ | $\alpha_3\left[\frac{Pa}{s}\right]$ |
|---|---|---|---|---|---|---|---|
| 100 | $2.4483 \cdot 10^7$ | $5.9452 \cdot 10^7$ | $5.6255 \cdot 10^6$ | $5.6535 \cdot 10^6$ | 7.4554 | 0.9950 | $1.3896 \cdot 10^{-4}$ |
| 200 | $1.7697 \cdot 10^7$ | $6.1442 \cdot 10^7$ | $5.5708 \cdot 10^6$ | $3.3737 \cdot 10^6$ | 2.7945 | 0.9832 | $1.1668 \cdot 10^{-4}$ |
| 300 | $1.2926 \cdot 10^7$ | $6.2529 \cdot 10^7$ | $5.9254 \cdot 10^6$ | $3.5228 \cdot 10^6$ | 6.3679 | 0.9947 | $1.5514 \cdot 10^{-4}$ |
| 400 | $6.1435 \cdot 10^6$ | $5.1562 \cdot 10^7$ | $9.5358 \cdot 10^6$ | $3.4654 \cdot 10^6$ | 9.2718 | 0.9912 | $1.0242 \cdot 10^{-4}$ |
| 500 | $5.3194 \cdot 10^6$ | $2.5746 \cdot 10^7$ | $1.2044 \cdot 10^7$ | $4.5884 \cdot 10^6$ | 9.9923 | 0.9172 | $1.2835 \cdot 10^{-4}$ |
| 600 | $6.1256 \cdot 10^6$ | $6.6388 \cdot 10^6$ | $5.1972 \cdot 10^7$ | $4.8862 \cdot 10^6$ | 10 | 0.0324 | $1.2531 \cdot 10^{-4}$ |
| 700 | $5.7461 \cdot 10^6$ | $6.2274 \cdot 10^6$ | $4.9139 \cdot 10^7$ | $6.1582 \cdot 10^6$ | 10 | 0.0256 | $1.4543 \cdot 10^{-4}$ |
| 800 | $5.6538 \cdot 10^6$ | $5.6422 \cdot 10^6$ | $4.5805 \cdot 10^7$ | $6.1727 \cdot 10^6$ | 10 | 0.0282 | $1.5695 \cdot 10^{-4}$ |
| 900 | $4.3561 \cdot 10^6$ | $7.2194 \cdot 10^6$ | $4.1366 \cdot 10^7$ | $6.5396 \cdot 10^6$ | 9.6898 | 0.0267 | $2.3905 \cdot 10^{-4}$ |
| 1000 | $5.0266 \cdot 10^6$ | $4.7903 \cdot 10^6$ | $3.8655 \cdot 10^7$ | $7.5268 \cdot 10^6$ | 10 | 0.0292 | $2.7670 \cdot 10^{-4}$ |
| 1100 | $4.6128 \cdot 10^6$ | $4.5402 \cdot 10^6$ | $4.5276 \cdot 10^7$ | $1.8722 \cdot 10^6$ | 10 | 0.0259 | $5.1502 \cdot 10^{-5}$ |
| 1200 | $4.1804 \cdot 10^6$ | $4.7878 \cdot 10^6$ | $4.4427 \cdot 10^7$ | $2.0907 \cdot 10^6$ | 10 | 0.0280 | $6.1143 \cdot 10^{-5}$ |

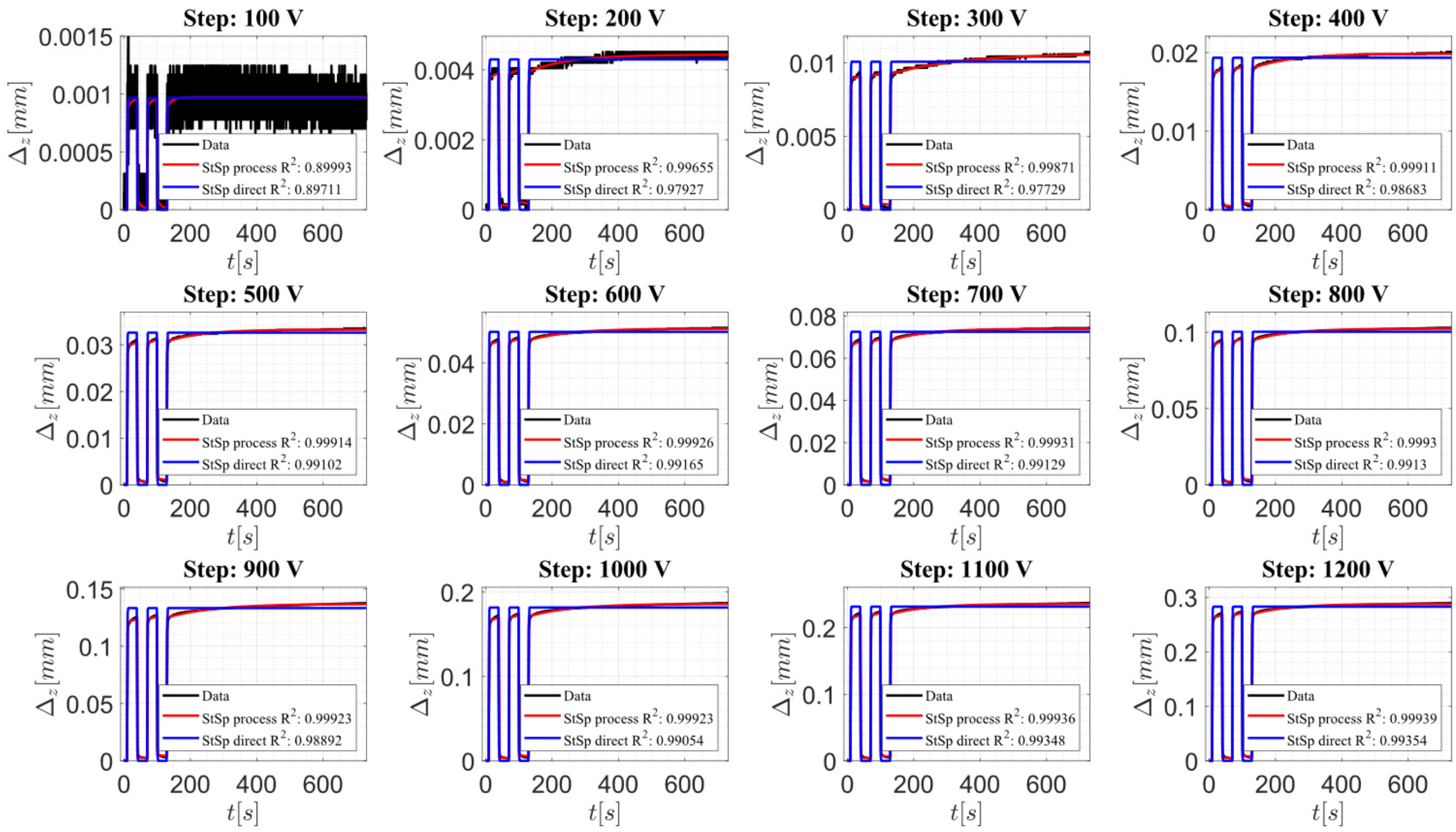

**Fig. S2.** Fit of third order discrete state space (StSp) model for all voltage levels

TABLE SVIII

IDENTIFIED PARAMETERS OF THIRD ORDER DISCRETE STATE SPACE MODEL

| Voltage [V] | A | B | C |
|---|---|---|---|
| 100 | $\begin{bmatrix} 0.9993 & 6.4406\cdot10^{-4} & 2.8609\cdot10^{-6} \\ 0 & 0.5089 & 0.0051 \\ 0 & 0 & 0.5089 \end{bmatrix}$ | $\begin{bmatrix} 3.4098\cdot10^{-7} \\ 0.0010 \\ 0.2327 \end{bmatrix}$ | $\begin{bmatrix} 5.0425 \\ 56.8986 \\ 0 \end{bmatrix}^T$ |
| 200 | $\begin{bmatrix} 0.9999 & 3.3094\cdot10^{-4} & 6.2901\cdot10^{-7} \\ 0 & 0.7738 & 0.0020 \\ 0 & 0 & 0.0169 \end{bmatrix}$ | $\begin{bmatrix} 1.7194\cdot10^{-7} \\ 0.0011 \\ 0.1542 \end{bmatrix}$ | $\begin{bmatrix} 2.6463 \\ 70.5301 \\ 0 \end{bmatrix}^T$ |
| 300 | $\begin{bmatrix} 1 & 2.6043\cdot10^{-4} & 5.8339\cdot10^{-7} \\ 0 & 0.7176 & 0.0023 \\ 0 & 0 & 0.0391 \end{bmatrix}$ | $\begin{bmatrix} 1.5615\cdot10^{-7} \\ 0.0012 \\ 0.1898 \end{bmatrix}$ | $\begin{bmatrix} 2.3209 \\ 75.8354 \\ 0 \end{bmatrix}^T$ |
| 400 | $\begin{bmatrix} 0.9999 & 2.9355\cdot10^{-4} & 8.5537\cdot10^{-7} \\ 0 & 0.6796 & 0.0033 \\ 0 & 0 & 0.1257 \end{bmatrix}$ | $\begin{bmatrix} 2.1805\cdot10^{-7} \\ 0.0015 \\ 0.2698 \end{bmatrix}$ | $\begin{bmatrix} 2.1903 \\ 61.8871 \\ 0 \end{bmatrix}^T$ |
| 500 | $\begin{bmatrix} 0.9999 & 2.7840\cdot10^{-4} & 8.5286\cdot10^{-7} \\ 0 & 0.6795 & 0.0035 \\ 0 & 0 & 0.1539 \end{bmatrix}$ | $\begin{bmatrix} 2.1496\cdot10^{-7} \\ 0.0016 \\ 0.2894 \end{bmatrix}$ | $\begin{bmatrix} 2.0592 \\ 61.3489 \\ 0 \end{bmatrix}^T$ |
| 600 | $\begin{bmatrix} 0.9999 & 2.5214\cdot10^{-4} & 7.6350\cdot10^{-7} \\ 0 & 0.6736 & 0.0035 \\ 0 & 0 & 0.1464 \end{bmatrix}$ | $\begin{bmatrix} 1.9316\cdot10^{-7} \\ 0.0016 \\ 0.2843 \end{bmatrix}$ | $\begin{bmatrix} 2.1173 \\ 69.3676 \\ 0 \end{bmatrix}^T$ |
| 700 | $\begin{bmatrix} 0.9999 & 2.7409\cdot10^{-4} & 8.8431\cdot10^{-7} \\ 0 & 0.6642 & 0.0038 \\ 0 & 0 & 0.1866 \end{bmatrix}$ | $\begin{bmatrix} 2.2087\cdot10^{-7} \\ 0.0017 \\ 0.3101 \end{bmatrix}$ | $\begin{bmatrix} 2.2137 \\ 66.2784 \\ 0 \end{bmatrix}^T$ |
| 800 | $\begin{bmatrix} 0.9999 & 2.7169\cdot10^{-4} & 8.0427\cdot10^{-7} \\ 0 & 0.6811 & 0.0034 \\ 0 & 0 & 0.1342 \end{bmatrix}$ | $\begin{bmatrix} 2.0424\cdot10^{-7} \\ 0.0016 \\ 0.2759 \end{bmatrix}$ | $\begin{bmatrix} 2.5810 \\ 78.8794 \\ 0 \end{bmatrix}^T$ |
| 900 | $\begin{bmatrix} 0.9999 & 2.6574\cdot10^{-4} & 9.1313\cdot10^{-7} \\ 0 & 0.7270 & 0.0044 \\ 0 & 0 & 0.2466 \end{bmatrix}$ | $\begin{bmatrix} 2.2219\cdot10^{-7} \\ 0.0019 \\ 0.3444 \end{bmatrix}$ | $\begin{bmatrix} 1.4773 \\ 47.5997 \\ 0 \end{bmatrix}^T$ |
| 1000 | $\begin{bmatrix} 0.9999 & 2.5498\cdot10^{-4} & 7.6745\cdot10^{-7} \\ 0 & 0.6927 & 0.0035 \\ 0 & 0 & 0.1446 \end{bmatrix}$ | $\begin{bmatrix} 1.9374\cdot10^{-7} \\ 0.0016 \\ 0.2831 \end{bmatrix}$ | $\begin{bmatrix} 2.5634 \\ 84.1382 \\ 0 \end{bmatrix}^T$ |
| 1100 | $\begin{bmatrix} 0.9999 & 2.4917\cdot10^{-4} & 7.6203\cdot10^{-7} \\ 0 & 0.6728 & 0.0035 \\ 0 & 0 & 0.1522 \end{bmatrix}$ | $\begin{bmatrix} 1.9238\cdot10^{-7} \\ 0.0016 \\ 0.2882 \end{bmatrix}$ | $\begin{bmatrix} 2.8483 \\ 94.3725 \\ 0 \end{bmatrix}^T$ |
| 1200 | $\begin{bmatrix} 0.9999 & 2.5217 \cdot 10^{-4} & 9.3562 \cdot 10^{-7} \\ 0 & 0.6929 & 0.0048 \\ 0 & 0 & 0.3183 \end{bmatrix}$ | $\begin{bmatrix} 2.2487 \cdot 10^{-7} \\ 0.0020 \\ 0.3811 \end{bmatrix}$ | $\begin{bmatrix} 1.6416 \\ 54.4909 \\ 0 \end{bmatrix}^T$ |

TABLE SIX
MODEL PARAMETERS NOT TO BE OPTIMIZED

| $z_0\ [m]$ | $\varepsilon_0 \left[\frac{s^4 \cdot A^2}{kg \cdot m^3}\right]$ | $\varepsilon_r\ [-]$ | $A_e\ [m^2]$ | $A_c\ [m^2]$ | $m\ [kg]$ | $j\ [-]$ |
|---|---|---|---|---|---|---|
| $25 \cdot 10^{-6}$ | $3.8542 \cdot 10^{-12}$ | 2.8 | $7.5602 \cdot 10^{-5}$ | $1.450 \cdot 10^{-4}$ | $3.4962 \cdot 10^{-6}$ | 399 |

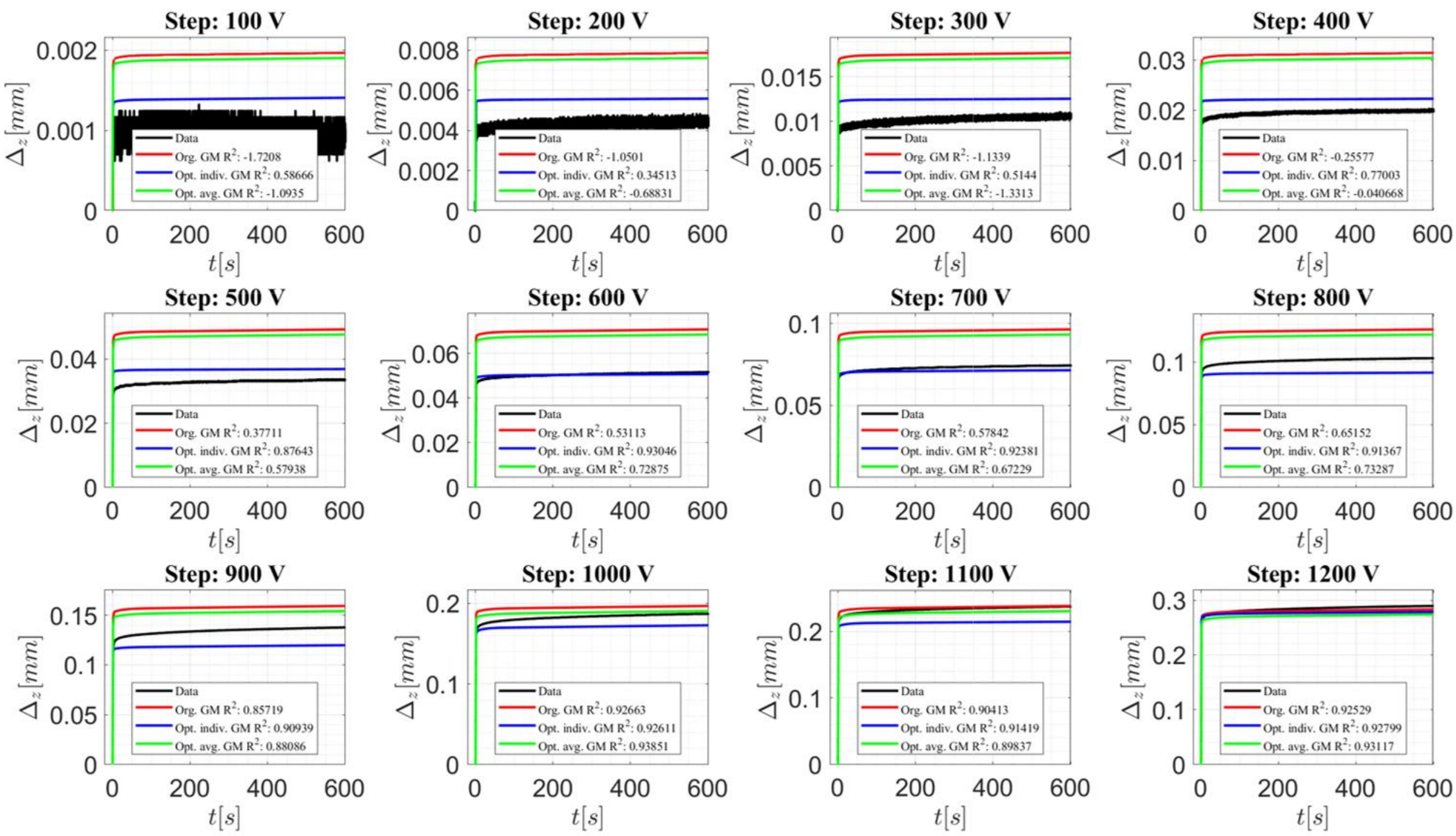

**Fig. S3.** Optimization results of generalized Maxwell (GM) model 1 with non- optimized (Org.), individual optimized (indiv.) as well as averaged and optimized parameters (avg.) for all voltage levels

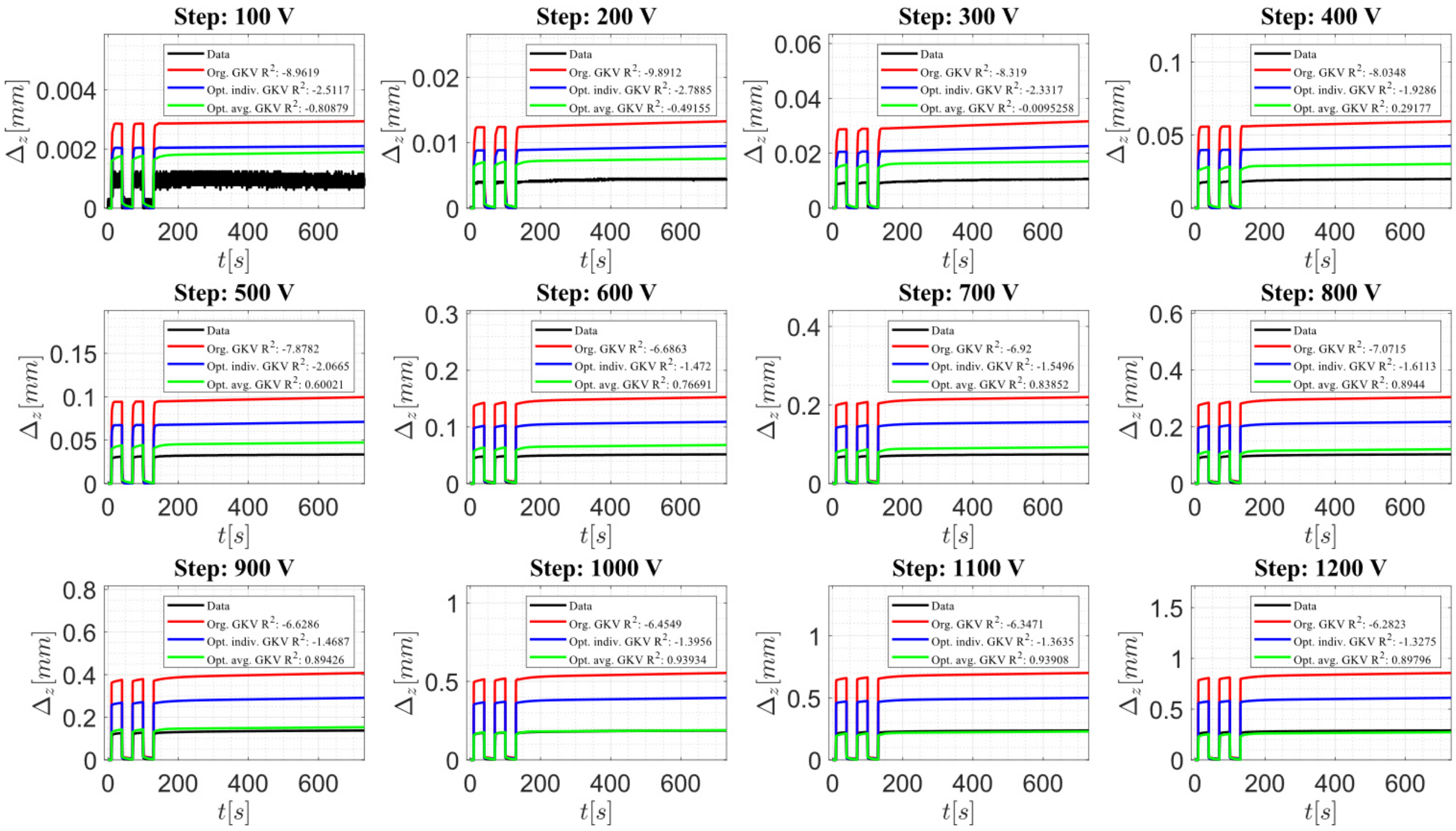

**Fig. S4.** Optimization results of generalized Kelvin-Voigt (GKV) model 2 with non- optimized (Org.), individual optimized (indiv.) as well as averaged and optimized parameters (avg.) for all voltage levels

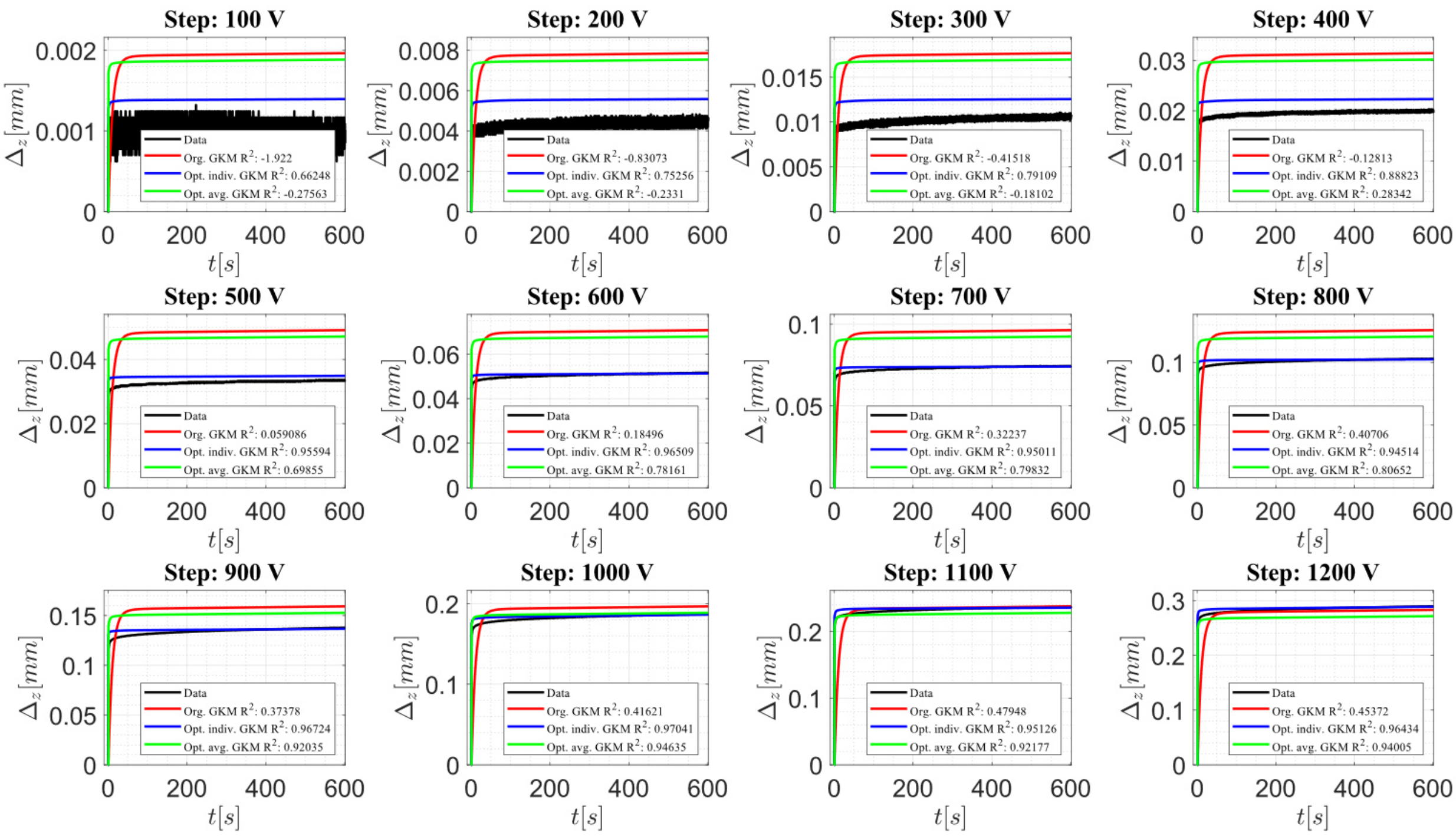

**Fig. S5.** Optimization results of generalized Kelvin-Maxwell (GKM) model 3 with non- optimized (Org.), individual optimized (indiv.) as well as averaged and optimized parameters (avg.) for all voltage levels

TABLE SX

ACTUATOR PARAMETERS FOR VALIDATION

| $z_0\ [m]$ | $\varepsilon_0\ \left[\frac{s^4 \cdot A^2}{kg \cdot m^3}\right]$ | $\varepsilon_r\ [-]$ | $A_e\ [m^2]$ | $A_c\ [m^2]$ | $m\ [kg]$ | $j\ [-]$ |
|---|---|---|---|---|---|---|
| $25 \cdot 10^{-6}$ | $3.8542 \cdot 10^{-12}$ | 2.8 | $7.5107 \cdot 10^{-5}$ | $1.1387 \cdot 10^{-4}$ | $3.4962 \cdot 10^{-6}$ | 399 |